# Multi-Oriented Text Detection and Verification in Video Frames and Scene Images

[a]Aneeshan Sain, [b]Ayan Kumar Bhunia, [c]Partha Pratim Roy*, [d]Umapada Pal

a. Dept. of EE, Institute of Engineering & Management, Kolkata, India
b. Dept. of ECE, Institute of Engineering & Management, Kolkata, India
c. Dept. of CSE, Indian Institute of Technology Roorkee, India
d. CVPR Unit, Indian Statistical Institute, Kolkata, India
*email: proy.fcs@iitr.ac.in, TEL: +91-1332-284816

## Abstract

In this paper, we bring forth a novel approach of video text detection using Fourier-Laplacian filtering in the frequency domain that includes a verification technique using Hidden Markov Model (HMM). The proposed approach deals with the text region appearing not only in horizontal or vertical directions, but also in any other oblique or curved orientation in the image. Until now only a few methods have been proposed that look into curved text detection in video frames, wherein lies our novelty. In our approach, we first apply Fourier-Laplacian transform on the image followed by an ideal Laplacian-Gaussian filtering. Thereafter K-means clustering is employed to obtain the asserted text areas depending on a maximum difference map. Next, the obtained connected components (CC) are skeletonized to distinguish various text strings. Complex components are disintegrated into simpler ones according to a junction removal algorithm followed by a concatenation performed on possible combination of the disjoint skeletons to obtain the corresponding text area. Finally these text hypotheses are verified using HMM-based text/non-text classification system. False positives are thus eliminated giving us a robust text detection performance. We have tested our framework in multi-oriented text lines in four scripts, namely, English, Chinese, Devanagari and Bengali, in video frames and scene texts. The results obtained show that proposed approach surpasses existing methods on text detection.

[1] Submitted to Neurocomputing, Elsevier

## 1. Introduction

With the advancement in technology, video-text detection and recognition has gained prime importance as real time applications, for instance, the systems that help blind people to travel on roads, or the ones that track license plates of automobiles, often require recognition accuracies greater than 90% [2, 3] for security and surveillance purposes. Due to motion-blur, non-uniform illumination, text movements and complex back-grounds [8] video images suffer significant loss in quality. Lighting conditions and perspective distortions [10], affects scene texts negatively as well. Thus, attaining such high accuracies is too idealistic an aim for researchers. Retrieval of textual information [31] from scene images or video frames has gained attention among the researchers because of text being one of the important medium of communication. Textual information retrieval system requires a robust text detection framework to localize the text accurately in scene images and video frames. In spite of many methods being proposed earlier, text detection remains a challenging problem owing to the unrestricted colors, fonts, and orientations of the characters.

Text detection methods may be categorized into three main approaches: connected component-based, edge-based, and texture-based [10, 12, 20]. The first method uses color quantization and region expansion (or splitting) to group adjacent pixels of similar colors into connected components (CC) [12]. As, these CCs may not retain the complete shape of the characters due to color bleeding and the low contrast of the text lines, these approaches do not hold good for video images. To mitigate this issue of low contrast, edge-based techniques are proposed. These methods [20] analyze horizontal and vertical profiles of the edge map. Albeit fast, these methods produce numerous false-positives for images having a complex background. To fix such problems, texture-based approach is applied which considers the text region as a unique contour or texture. These techniques employ Fast Fourier Transform, discrete cosine transform, wavelet decomposition, and Gabor filters for extracting features. Such methods usually apply classifiers, for instance SVM and neural networks. Hence they need to be trained for different databases. But, these classifiers need an enormous training set of text and non-text samples for higher accuracy [10].

Multi-oriented detection of text without any restrictions on background, alignment, and contrast, with high precision, and recall still remains a difficult task. Most existing methods [5, 9, 19] depending highly on the horizontal text-orientation fail in cases of multi-oriented text fields. Research on curved-text-line detection is even rarer because not only edge-focused methods fail but also due to most of the above mentioned reasons. Hence, in this paper, we introduce a method which can handle linear texts of arbitrary orientation, as well as curved text lines. In this paper we further propose HMM-based text verification for higher accuracy.

The proposed text detection method moves forward from baseline approach [1] with improved performance. We have used Laplacian of Gaussian filter for better filtering. The proposed system takes care of curved texts as well as texts in any orientation. One of the novelties lies in using skeletal features extensively in detection of curved texts which may be in any orientation. In addition, HMM based verification has been included to enhance the efficiency and accuracy of the framework.

The contributions of the paper are as follows – 1. An efficient approach is proposed which is able to detect horizontal, non-horizontal and curvedly oriented texts in videos. 2. The concept of skeletonization is extensively used to improve the detection process of text region. 3. HMM verification is applied to improve accuracy of results. 4. Finally, the framework has been tested with 4 different scripts to show the efficiency.

The rest of the paper is organized as follows. In Section 2 we discuss related work developed for scene/video image text detection. In Section 3, the proposed framework for text detection and verification has been described. In Section 4 we detail the experimental setup and discuss about the results. Finally, conclusions and future directions are given in Section 5.

## 2. Related Work

Text detection has been an appealing topic of exploration to most researchers. A considerable amount of work has been done in text detection in scene and video images [2-6]. Two recent survey papers on text detection and recognition in scene images and video frames can be found in [2, 47]. Edge based methods are common in application. Cai et al. [20] designs two filters for enhancing the edges in text regions using different threshold values to decide whether or not to do so in a specific region. Hence it does not generalize well for various data sets. Liu et al. [9]

uses the Sobel edge maps of four directions, extracting statistical features from them and classifies pixels into text and non-text clusters applying K-means algorithm. Although robust against complex backgrounds, this method cannot locate low contrast text. Moreover it is computationally expensive owing to the enormous feature set.

Rong et al. [26] proposed a two-level algorithm to detect text regions in natural scene images based on the characteristics of character components. Chen et al. [27] proposed a method for robust text detection in natural scene images using Maximally Stable Extremal Regions (MSER) and stroke width distance based features. Yin et al. [24] suggested robust text detection in natural scene images based on MSER, clustering and character classifier. The method studies the characteristics of character components for the output of MSER to classify them as text candidates. Single link clustering and character classifiers are used for detecting true text candidates. Since the above discussed methods assumes that a given image has high contrast, the methods like MSER outputs character components. However, if the same methods deployed on video image, the performance of the method degrades severely due to low contrast and low resolution. Therefore, these methods are sensitive to complex back ground, distortions and low contrast.

In video images, the performance of MSER falls severely owing to its low contrast and low resolution. Texture based approaches were also used by many researchers. Shivakumara et al. [1, 21] proposed wavelet, color features, Fourier with color spaces and Fourier with Laplacian for text detection in video. These three methods are good for low contrast images but are computationally expensive since they use expensive transformation. In addition, the performances of the methods degrade when an input image contains distortion caused by operations and motions. Lee et al. [6] used support vector machines (SVM) to categorize each pixel into text and non-text classes. But again, texture-based methods tend to use large training sets which are computationally exhausting.

Other mechanisms and methods implement gradient and edge information in various ways to improvise their performance. Epshtein et al. [5] proposed canny edge images and stroke width transform (SWT) to detect text in natural scene images. Roy et al. [4] suggested an approach for identifying texts using binarization which is based on the concept of fusion. A blind deconvolutional model [50] was used for text detection by enhancing the edge information of the text regions. Here, classification between blurred and deblurred images was performed and it was followed by a deblurring operation which used Gaussian weighted-L1 for restoring sharpness of the edges in blurred images. Recently, deep learning based methods are being explored in various scene and video text detection and recognition works [7, 48, 49, 55]. A Text-Attentional Convolutional Neural Network (Text-CNN) has been used in [48]. In this work, the authors used rich supervised information related to text region mask, character label, and binary text/non-text information to train the network which increases the robustness against complicated background component. Cho et al. [53] proposed a text detection system, called Canny Text Detector which takes advantage of the similarity between image edge and text regions for robust text localization. Although this method works satisfactorily for video texts, arbitrary orientations are not dealt with. Other works associated with scene text detection in camera-based images can be noticed in [22, 23] (that focus majorly on text orientation). Although, these methods perform well for camera-based images, they are not efficient for video images as high resolution text is required for clear shape detection and extraction of each character [15]. In [43] text localization in natural scene images is performed by edge recombining, edge filtering and multi-channel processing sequentially. Confidence map and context information have been explored in [44] for text detection in scene images. A multistage clustering algorithm for grouping MSER components to detect multi-oriented text was proposed in [51]. Liang et al. [52] proposed a method based on multiple-spectral fusion for arbitrarily-oriented text detection in video images. They introduced an idea of convolving Laplacian with wavelet sub-bands in the frequency domain to enhance the low resolution text pixels. Very recently, a Ring Radius Transform (RRT) based text detection method [54] has been proposed for in multi-oriented and multi-script environment. A Fully Convolutional Network (FCN) model based multi-oriented text detection framework was introduced in [55] which considered both local and global cues into account in a course-to-fine procedure. FCN model was trained to predict the salient map of the text which was combined with the character components. Finally, another FCN classifier was used in order to remove the false hypothesis. In [60], Histogram of Oriented Moment feature is used for text/non text classification using Support Vector Machine (SVM). Authors used a sub pixel mapping based super resolution approach to enhance the image before text detection. Shivakumara et al. [29] fractal properties have been used in the gradient domain to enhance low resolution mobile video, which has been further explored into the wavelet domain in a pyramid structure for text components in text clusters to make the method font size independent. Related works in curve text detection [13] are less due to its complexity. In this proposed work we have explored skeletonization on differently oriented texts of varying sizes and shapes, along with HMM based verification for robust multi-oriented text detection system.

## 3. Proposed Framework

The approach followed in this paper can be broadly divided into two processes; text detection, and text verification. The text detection process consists of two major steps. First, we identify candidate text regions using Fourier-Laplacian filtering; thereafter applying a maximum-difference map to segregate the image into text and non-text regions. Secondly, the text regions obtained from the first step are reduced to their skeletons. These skeletal figures are analyzed and their branch-points are separated to provide individual branchless continuous skeletons. For extraction of curved text regions we have used curve-fitting methods that use polynomial equations as described later. The chosen text hypotheses are verified through HMM and the wrongly chosen text areas are neglected thus eliminating false positives. Finally after verification from HMM the output is obtained. The flowchart in Fig.1 shows the system in a nutshell. These steps are detailed as follows.

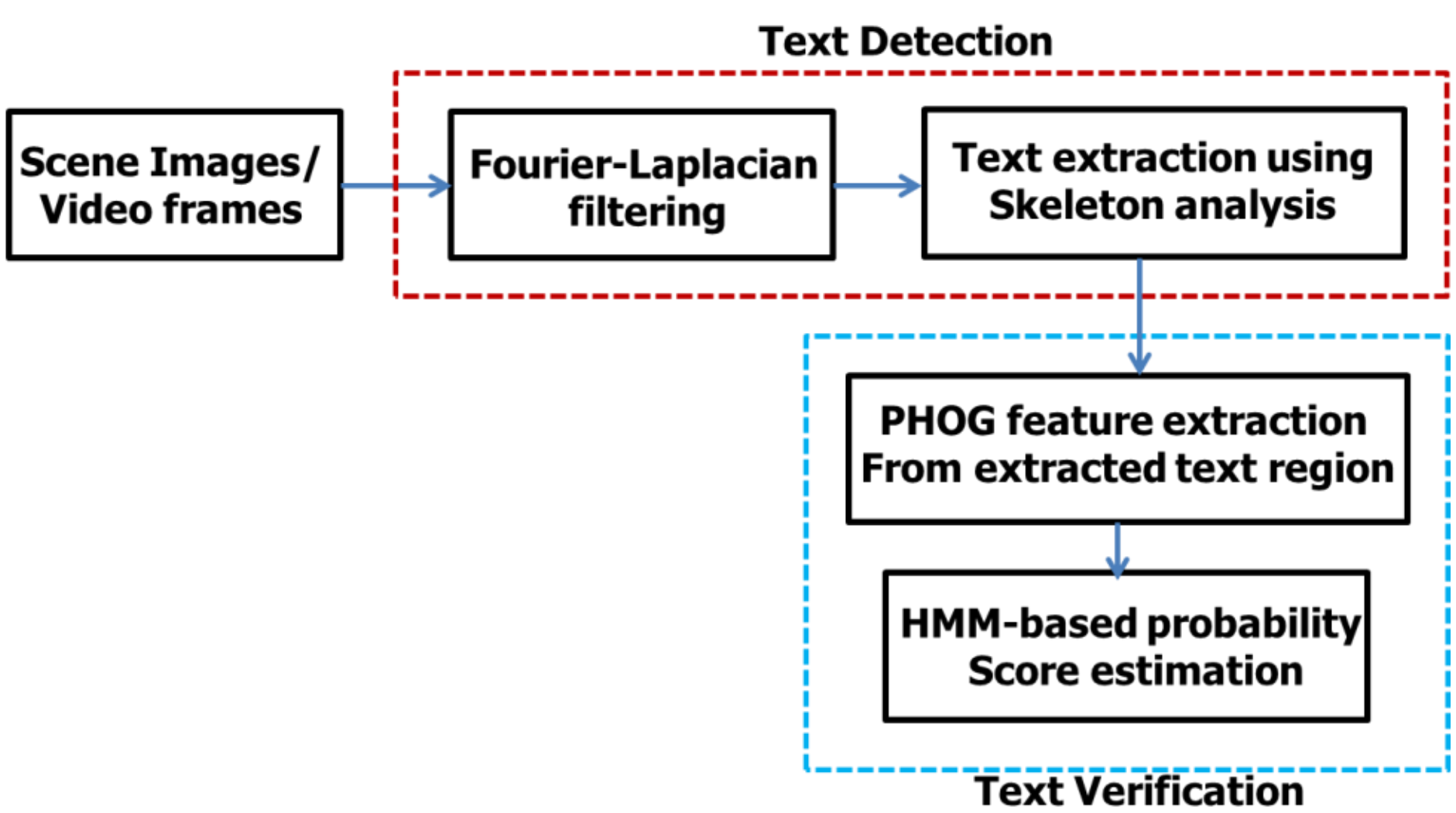

**Fig.1. Flowchart of our proposed approach**

## 3.1. Text Area Detection

In this section we describe the processes involved for detection of the candidate text regions in the image. Firstly, Fourier-Laplacian filtering pre-processes the image for text detection. It highlights the probable text regions which is further processed and skeletonized thereafter. The skeletons of the possible text portions of the image are meticulously nurtured and processed to perfect the accuracy of the chosen candidate text regions. Then the corresponding text region is extracted from the image and sent for verification. The block-diagram below (In Fig 2) shows the steps involved in this process, and the processes are described in detail thereafter.

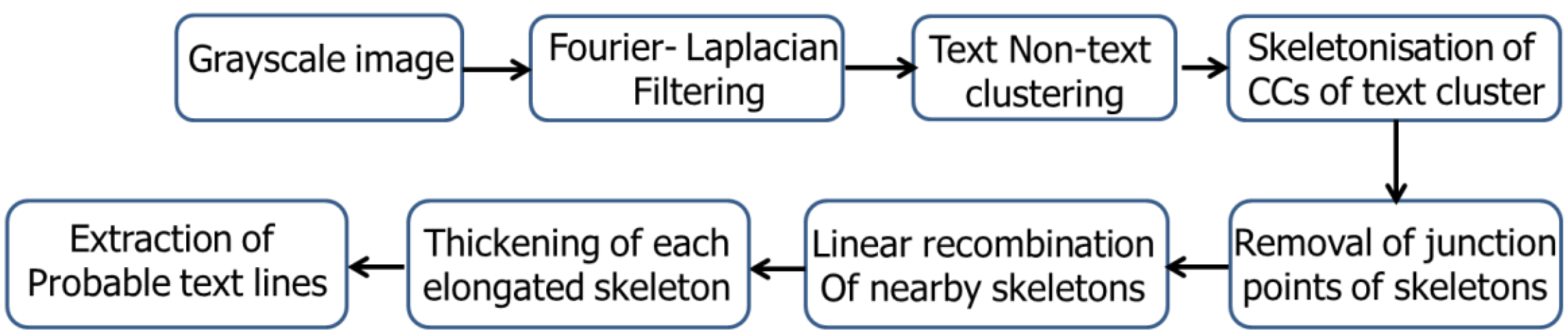

**Fig.2. Block-diagram describing the text detection processes**

### 3.1.1 Fourier-Laplacian Approach to Text Detection

Video texts can have very low contrast against complex local backgrounds. It is necessary to preprocess the input image to distinguish between text and non-text areas. Here we apply Fourier-Laplacian filtering to smoothen noise and detect probable text regions as discussed in Shivakumara et al. [1]. Firstly, Fast Fourier-Transform is applied on the greyscale version of the input image to bring it into the frequency domain. Then an ideal low-pass filter is used in the frequency domain to smoothen noise. The motivation behind this step is to remove the high-frequency components of the Fourier transform, which contains information about noise. Thereafter we use Laplacian of Gaussian (LoG) mask in the frequency domain to detect probable text areas (Fig. 3(a) and Fig. 3(b)).

It is observed that, text regions contain many positive and negative peaks in intensity values, while non-text regions do not. It is also noted that the zero crossings refer to the transitions between text and background. Ideally, there will be equal number of text-to-background and background-to-text transitions. However, this case does not hold true for low contrast text on complex background, which is why we apply weaker conditions to make sure that low contrast text is captured as well.

Here, we have stressed on Fourier-Laplacian filtering, as it is more efficient than Laplacian filtering alone, owing to the fact that the former method of filtering aids in smoothening noise in video images thus decreasing the undesired noise sensitivity issue of the Laplacian operator. Fig.3(a) shows an input image and its Fourier-Laplacian filtered version in Fig 3(b).

It is noteworthy that, LoG filtered image (Fig. 3(d)) contains more text pixels than the Laplacian filtered one (Fig. 3(c)). So, it can be concluded that the LoG filter takes into consideration low contrast text information which is beneficial at a later stage for text extraction and thus we use LoG mask in frequency domain for subsequent processing of the image.

The next step uses a Maximum Difference (MD) map [14] on the output of Fourier LoG filtered image. The MD map of Fig 3(e) is obtained by using a non-overlapping sliding window of 1xN pixel units to scan through the image shown in Fig 3(e), where N is calculated as:

$$\mathrm{N} = \max(\mathrm{height}, \mathrm{width})/20 \quad \ldots\ldots\ldots\ldots (1)$$

For every window placed on the image, the intensity values of all the pixels in that particular window are replaced by the difference of the maximum and the minimum pixel intensity values in that window of the image.

The MD map thus obtained can be seen in Fig. 3(e). Thereafter, based on the Euclidean distance of MD values, K-means clustering method is applied to segregate all the pixels from the MD map into two clusters, ‘text’ and ‘non-text’. The pixels having lower intensity value are clustered as non-text and those pixels having higher intensity values are clustered as text. The morphological operation of ‘opening’ clears away small artifacts (see Fig. 3(f)). This results a reduced scope for search of actual text regions by eliminating non-text regions. After obtaining these refined areas for probable text regions, the processes involved in text extraction are applied which are discussed in details in the next section.

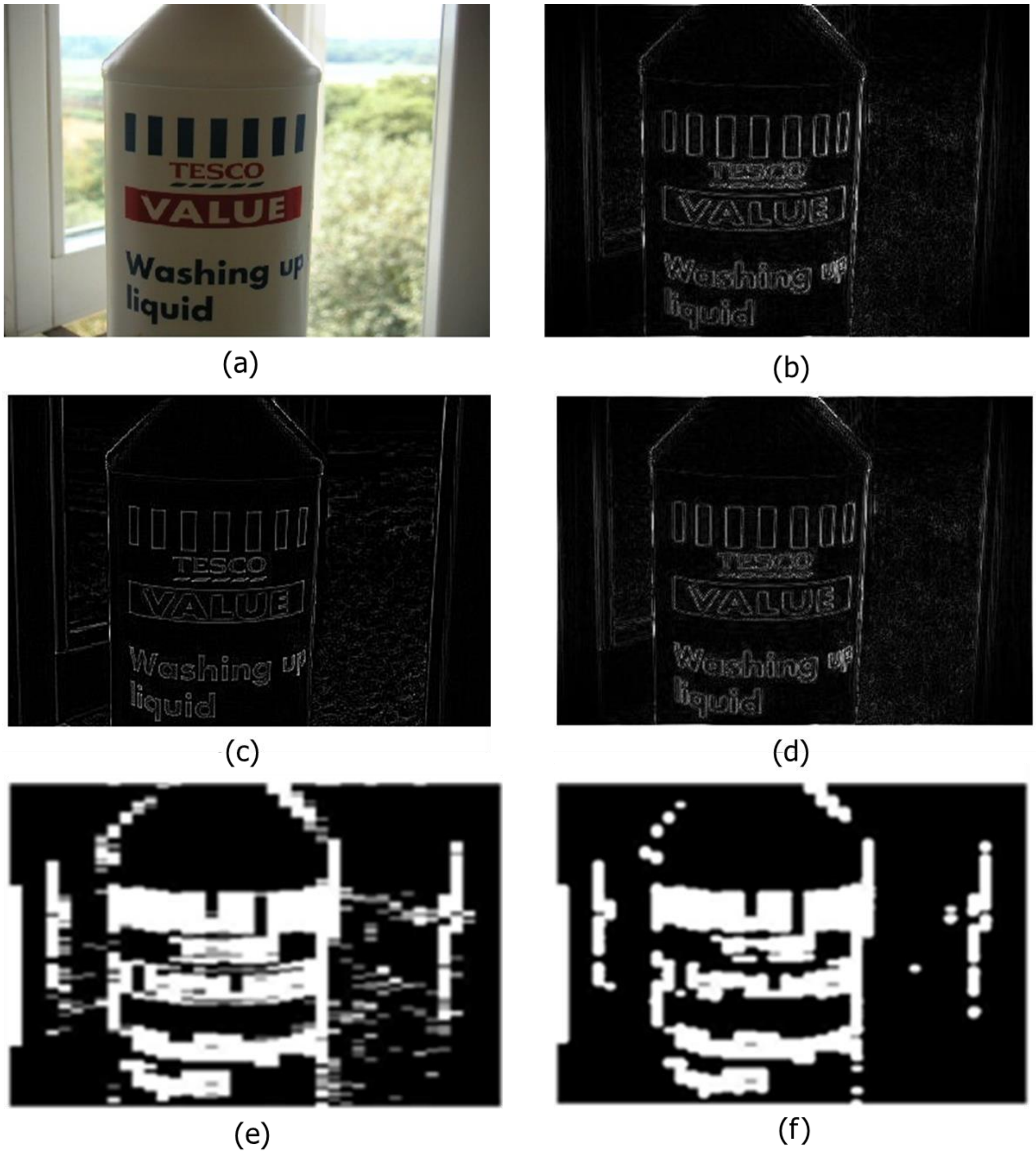

**Fig.3. (a) Original Image (b) Fourier-Laplacian filtered (c) Applying Laplacian filtering in Fourier domain (d) Applying Laplacian of Gaussian(LoG) filter in Fourier domain. (e) MD map of image of (d). (f) Text cluster after morphological opening of (e).**

### 3.1.2. Text Extraction

This section deals with the processes that would extract the candidate text region from the original image and send it for HMM-based verification (explained in next section). The objective behind these processes is the proper skeleton analysis of the probable text patches obtained from the previous steps. The features of skeletons; such as, shape, length, branching etc. play a crucial role in our text-detection process.

The connected components (mostly text regions) that are obtained after clustering the MD map information are analyzed next to remove false positives. Usually, bounding boxes can be employed to detect these text blocks [34]. This shows satisfactory results for horizontal text lines. However for curved text, or for text-lines in any other orientation, rectangular boxes will also enclose excess background pixels. Text lines that lie in close vicinity, might also be included, thus leading to overlapped bounding boxes. Hence we use ‘skeletonization’ to reduce every CC into its skeletal form (see Fig. 4(a)). Skeletonization [35] has been applied as a coarse method of classifying image regions as text regions or non-text regions. It provides an approximated region of text along with the direction alignment of probable candidate text region after the Fourier-LOG filtering process. Another reason for skeletonisation is to avoid the usage of bounding boxes, which would yield many false positives, when the text regions are curved.

Junction points are a fundamental part of skeleton analysis [35]. They refer to the points on a skeleton where branching takes place. The junction point is calculated as follows. Considering a pixel 'P' on the skeleton we count the number of connecting pixels surrounding it in a 3x3 window, centering the pixel 'P'. If there is only 1 connecting pixel, we conclude it to be an ‘end-point’. If there are two such connecting pixels, it is just another point on the skeletal path, no branching occurs. If there are more than 2 connecting points we conclude the point to be a junction or branch-point.

While dealing with a skeleton component, the co-ordinates of all the points corresponding to it are stored, and thereafter the points of each skeleton-branch are traced to calculate distance. The skeletal segment having longest distance (the distance between the two farthest placed endpoints) is considered as the main axis. The large skeleton branches having length greater than 1/3$^{rd}$ of the length of the main skeleton are separated by removing those junction points from the main skeleton. Remaining branches are having short length and hence their junction points were not removed. Those small branches are required for width calculation of image patches later on.

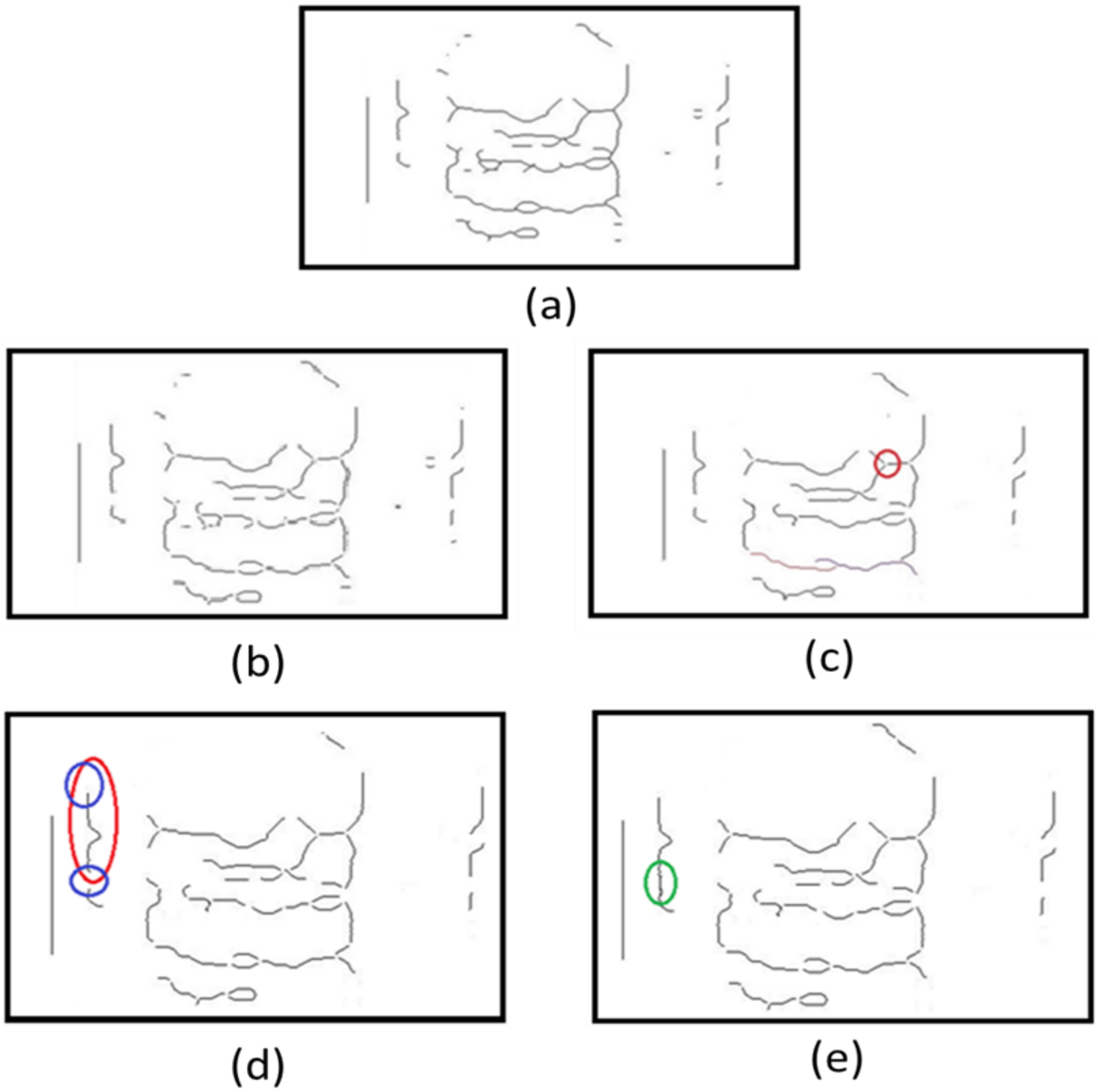

**Fig.4. [Color-inverted for better visibility] (a) Skeletonized results of Fig(3f). (b) Junction points are removed and joints are separated. (c) Small artifacts from (b) are cleared away. Red circle shows an area possible for curve text. (d) Same as (c). Red ellipse shows the primary skeleton in focus and the two blue ones indicate the allowed vicinity to search for other skeletal segments to join. (e) Skeletal segment obtained after joining the two skeletons. The green circle shows the area where each of the individual skeletal segments has been extended to join.**

The image thus obtained is now scanned and every encountered skeleton is considered. Note that this image may have more skeletal segments, but the skeletons are simpler and do not have long branches as they have already been separated. Thereafter a threshold value (T) is set based on the average length of all newly formed skeletons (Here, $T = 1/7^{th}$ of the average length of all skeletons) and remove all such spurious fragments having length less than T. Thus unwanted tiny artifacts that might wrongly interfere with the successive operations are cleared away (Fig. 4(c)).

Now, for each skeletal segment encountered, we search for another end-point of other skeletal segment in its vicinity (which is equal to a radial distance of $1/10^{th}$ the length of that skeletal segment), to join with it (Fig. 4(d) and Fig. 4(e)). It is due to the fact that longer text-lines provide better information for text/non-text verification through HMM process. The newly found skeleton in that scope is joined with the skeleton under consideration by extending both towards each other. Next, the skeleton component is thickened to a certain width. To find the width up to which the skeleton should be thickened, the endpoints of the skeleton are considered. To explain this thickening method, we show an example in Fig. 5(a) that represents a simple text patch. A rectangular block is considered for simplicity. Looking at the endpoints of the branches that arose from the main vertical skeleton (denoted as XY in Fig. 5(b)), we observe that the distance between the endpoints, that are symmetrically placed about the main skeleton, (e.g. A and B) would give us the approximate width of the text patch.

The width of each branch (of skeleton) may be different and is calculated as the distance between endpoints placed symmetrically about the main axis of the skeleton. Note that, symmetrically placed endpoints refer to the endpoints of the short side branches which lie approximately in equal distance on opposite sides of the main axis. To find this symmetric behavior, first, on encountering a skeletal segment, co-ordinates of the skeleton points are stored. Next we choose the main skeletal axis, as defined above. Thereafter we note the distances of the end-points of the branches calculated from the main skeleton axis. Two points are considered symmetric if (a) their distances from the main skeleton are approximately equal, and (b) the sum of distances of those points from the main axis is equal to the distance between those two points. Let two points be 'a' and 'b'; their distance from the main axis be $d_a$ and $d_b$ respectively; and the distance between them be $d_{ab}$. Here '$d_a$' is calculated as the Euclidean distance between the endpoint 'a' and the junction point on the main axis from where the branch starts. Similarly, '$d_b$' is calculated. The Euclidean distance (d) between two points having co-ordinates $(x_1, y_1)$ and $(x_2, y_2)$ is given by:

$$d = \sqrt{(x_1 - x_2)^2 + (y_1 - y_2)^2} \text{ ……….. (2)}$$

Now these two points 'a' and 'b' will be symmetrical if,

$$\left(\frac{|\,d_a - d_b\,|}{\max(d_a,d_b)} < 0.05 * \max(d_a, d_b)\right) \;\&\&\; \left(\frac{|\,(d_a + d_b) - d_{ab}\,|}{\max((d_a + d_b), d_{ab})} < 0.05 * \max((d_a + d_b), d_{ab})\right) == \text{TRUE}$$

……….. (3)

The maximum of all such '$d_{ab}$' distances is chosen as the width (W) of the skeletal segment. During symmetric end point computation, it may happen that some branches do not have a symmetrical component, such as in Fig. 6(b). In this case, '$d_{ab}$' is calculated as twice the distance of that endpoint of that branch from the main skeletal axis. Also, in the case where there are no branches in the skeletal segment, we set the value of width (W) to 1/3$^{rd}$ the length of the skeletal segment.

Thereafter we thicken the candidate text skeleton to a width of W units, thus creating a white patch signifying the indicated text region (like the text patch in Fig. 5a). Next, from the main grayscale image, we extract the text area corresponding to the white patch obtained, to get the required text line, which is thereafter passed to HMM-based verification.

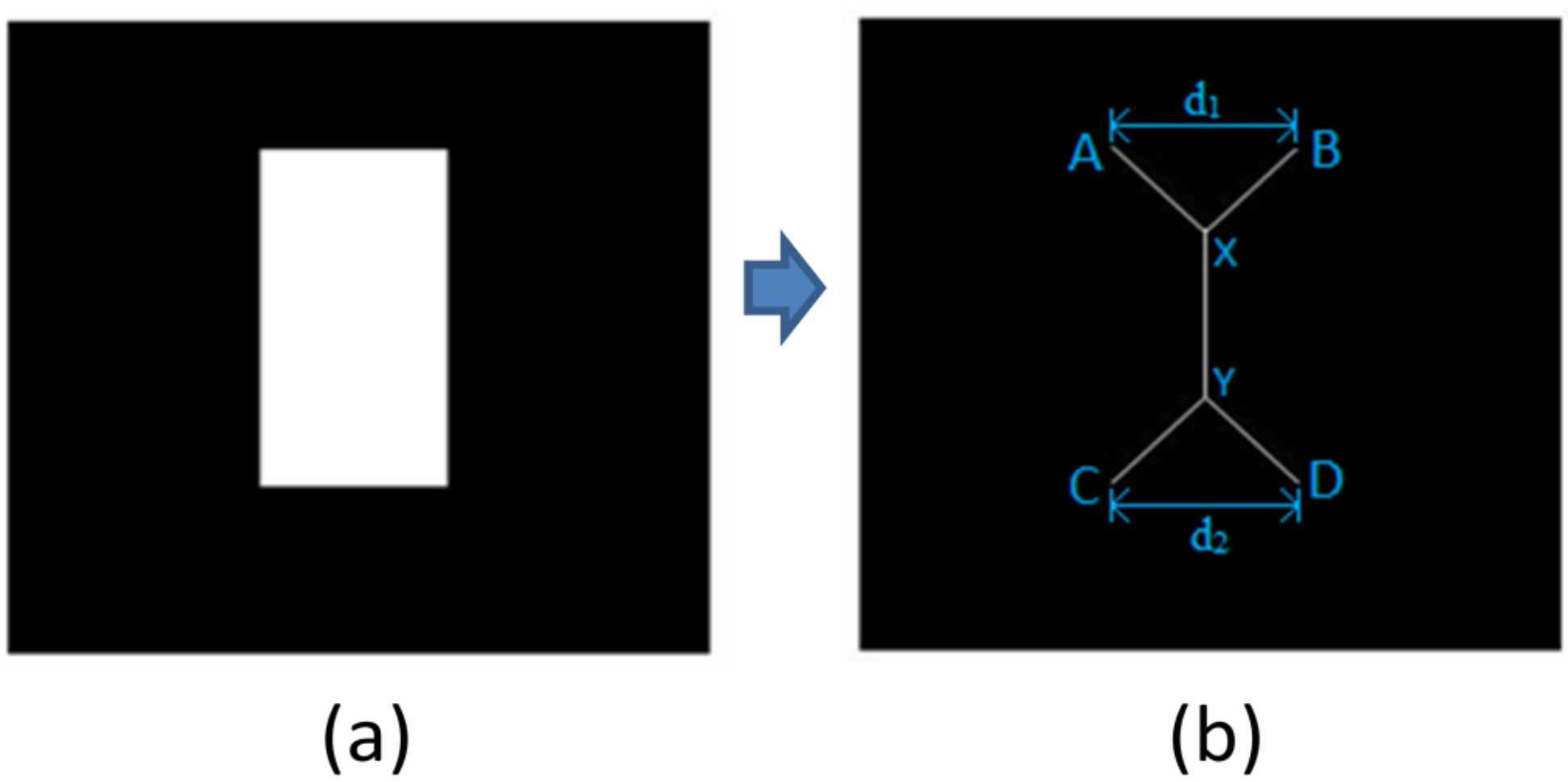

**Fig.5. (a) An example of simple white patch. (b) Its skeleton (in white).**

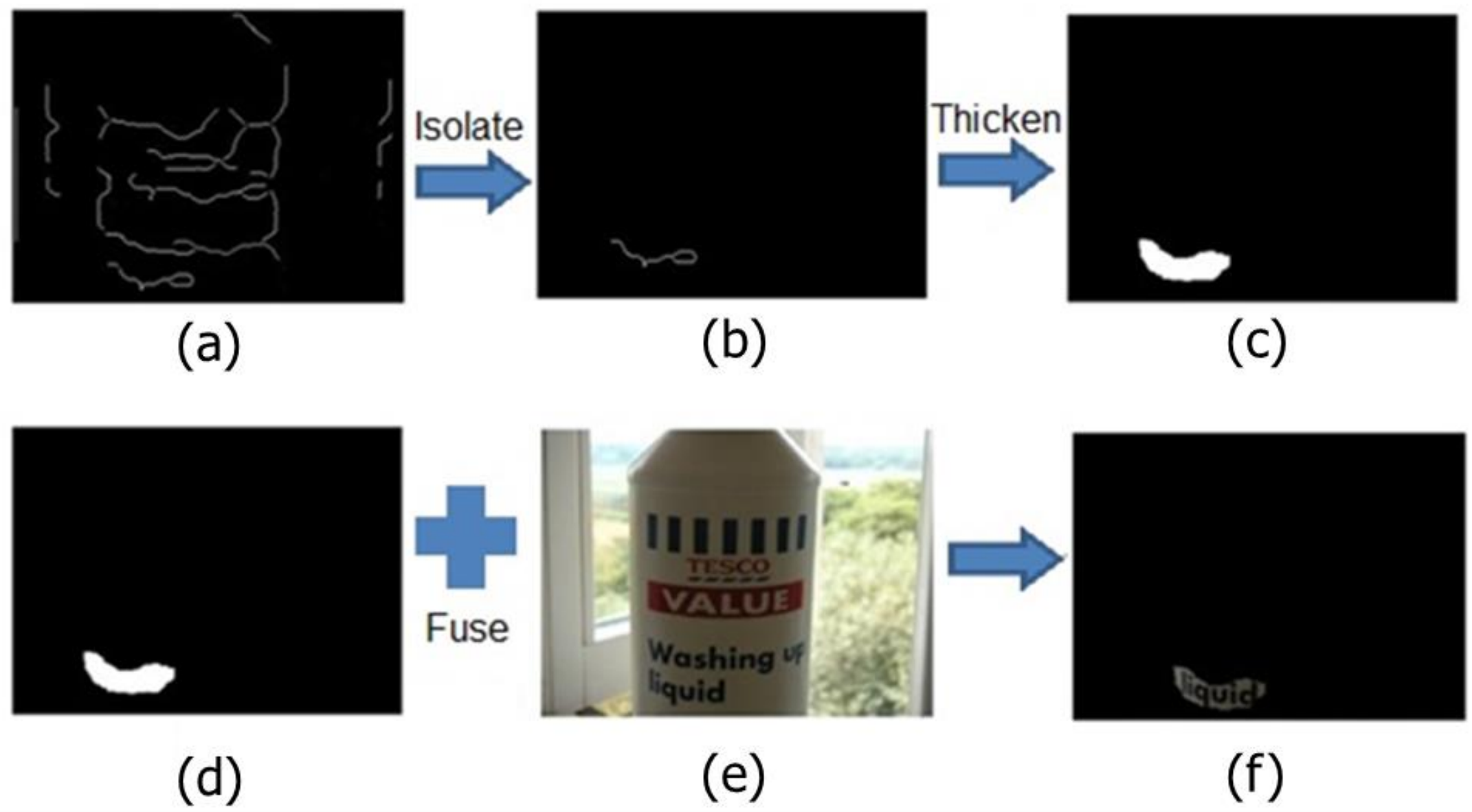

**Fig.6. Text extraction show for only 1 such skeletal segment. (a)Skeletonized image of Fig 3(b). (b) One particular skeleton is considered. (c) That segment is thickened to a width W. (d) same as (c). (e) Input image. (f) Text region having the shape of the white patch in (c) extracted out.**

If it is verified as text-region, the extracted text position is used as a mask to get the region from grayscale of the original image; else the skeleton is discarded permanently. The process discussed so far has been described pictorially in Fig. 6.

It may happen that there will be a situation like the one shown in Fig. 7, i.e. a detached junction point in the complex connected segment where there are more than multiple skeletal segments in the vicinity of our selected skeleton's end-point, all suitable for joining, (junctions shown in red circles). For this purpose we apply the following procedure. We remove only those junction points whose branches have a length greater than $1/3^{rd}$ of the length of the main skeleton. Note that, all available junction points are not removed as mentioned earlier in Section 3.1.2. As is evident from Fig.7(c), all the other junction points apart from the ones shown in red, have branches shorter than the specified length. Those small branches are required for width calculation of image patches later on. So, in this case, we observe, that all the other skeletons are of considerably similar lengths, thus determining it to be a skeleton of a curved text. In Fig 8, the three skeletons marked '1', '2', and '3' are in close vicinity about their junction point (shown as a blue circle).

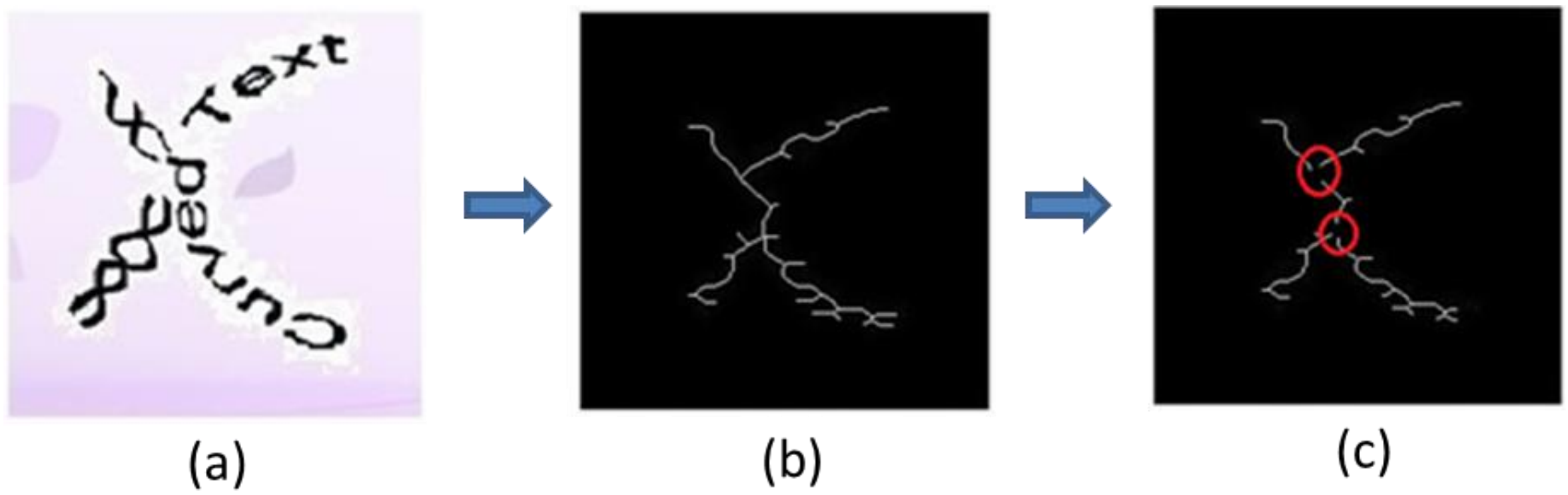

**Fig.7. (a) Original Image. (b) Image is skeletonized and small artifacts removed. (c) Junction points removed. The red circles indicate the junction points that have multiple skeletal segments starting in close vicinity.**

Each of these skeletons is thickened in the same way as described earlier, with respect to Fig. 5. These thickened regions are added to a blank image as a white patch separately (Fig. 8). Thereafter the text area corresponding to the white text patch is extracted from the grayscale version of the main image and sent for HMM-based verification (explained in the next section). The same process is followed for the other skeletal segments in the junction scope (i.e. skeletons marked '2' and '3'). We discuss HMM-based text verification in the following section.

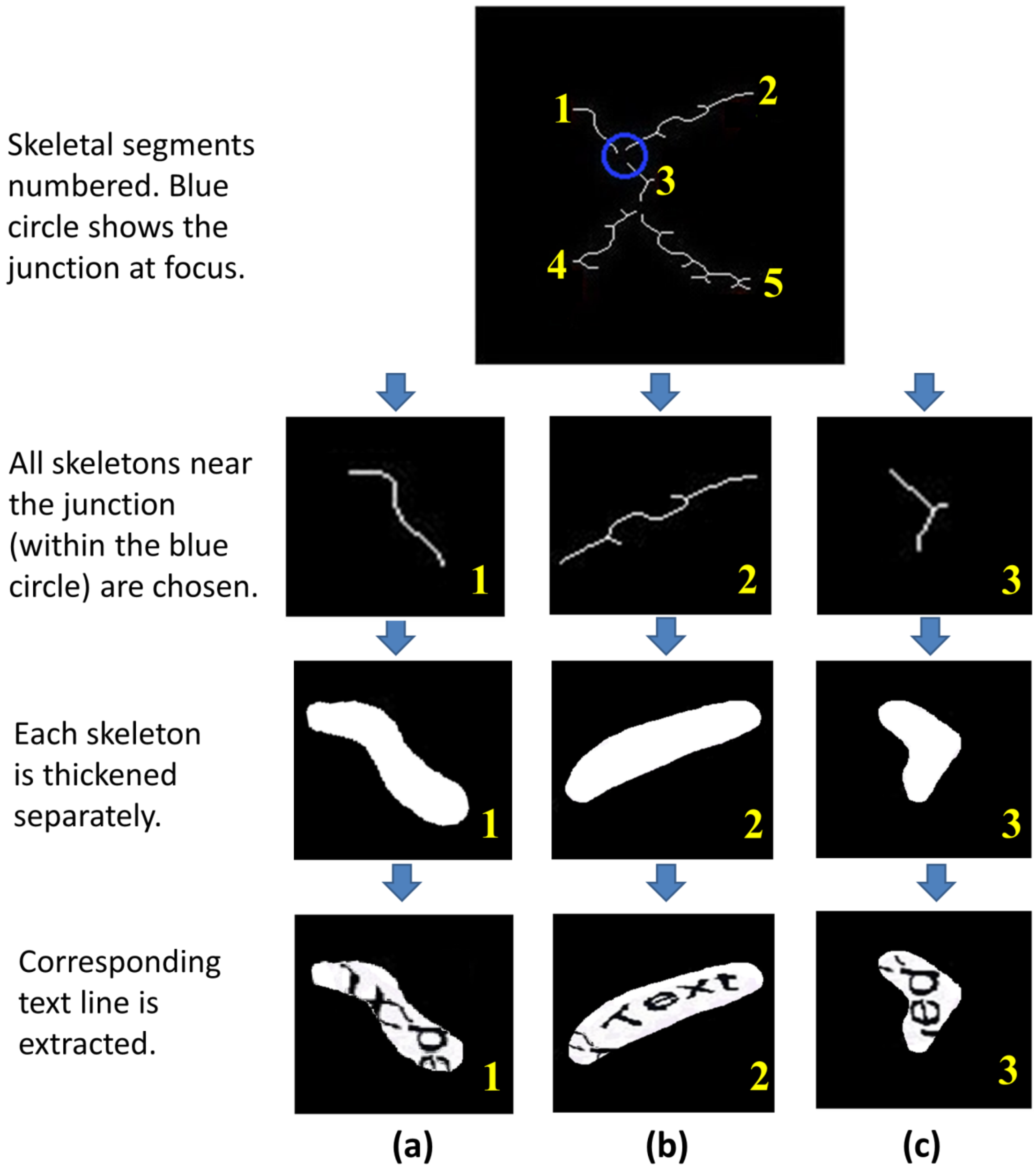

**Fig.8. Text extraction process for a curved text. (a),(b),(c) : Different possible text-lines out of which 2 when joined, will represent the curved text (focusing on the junction marked in blue). In this case, they are skeletal segments (2) and (3).**

## 3.2. HMM based Text-Verification

From the experiments performed, we have observed that there are cases where previous false positive elimination methods fail to identify non-text regions in the image. To handle these kinds of scenarios, we propose a HMM–based text verification approach to retrieve the proper text region. For this purpose, first, we use log-likelihood score estimation based on HMM to get the properly oriented skeleton of multi-oriented text region as well as to handle the case of curved text regions. Next, we verify the segmented text line using HMM based classification. The following flow chart (see Fig. 9) describes the chronological order of the steps involved.

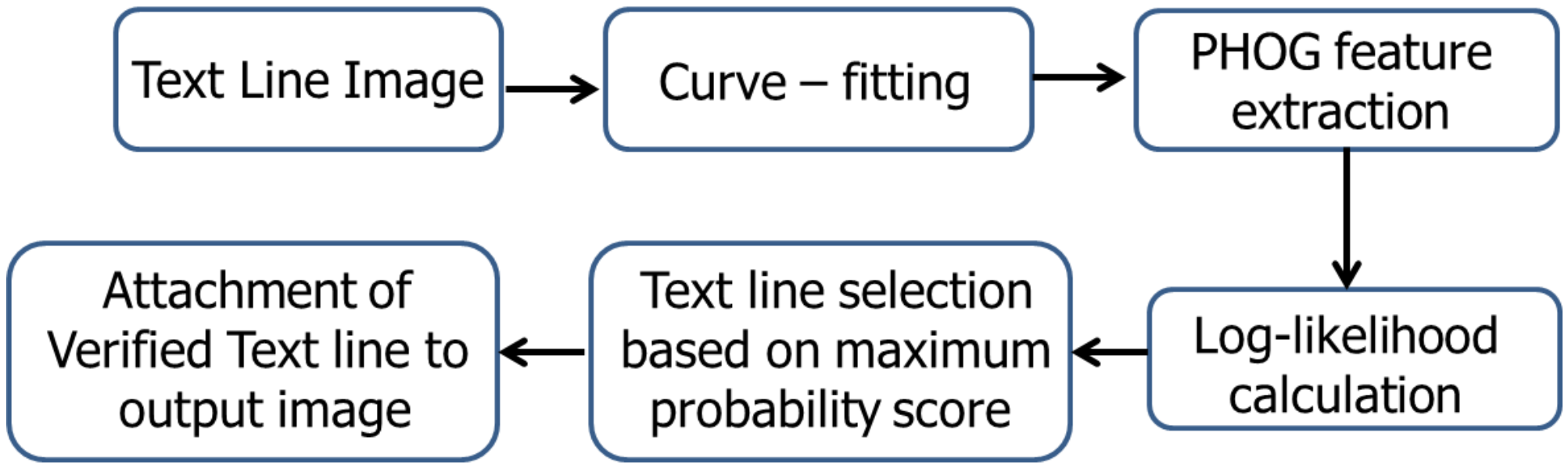

**Fig.9. Flowchart describing the text verification process**

### 3.2.1. Feature for HMM

We have used sliding window based PHOG feature for feature extraction purpose. A sliding window of width 40×8 is being shifted from left to right with 50% overlapping between two sliding window positions and gradient feature is calculated for each sliding window position.

To take care of multi oriented text, this movement of sliding window should follow the polynomial equation fitted for the multi-oriented word image as mentioned in [18]. A general polynomial equation is given by:

$$f(x) = a_n x^n + a_{n-1} x^{n-1} + \dots\dots\dots + a_1 x^1 + a_0 \quad \dots\dots\dots\dots (4)$$

Where $n > 0$ and $a_0, a_1 \dots. a_n$ are positive real numbers which are to be evaluated from the curve fitting algorithm. We use a fourth degree polynomial, using n = 4. The height of the image-patch is found taking the average of the height from each point of the polynomial equation.

PHOG [16] is the spatial shape descriptor which gives the feature of the image by spatial layout and local shape, comprising of gradient orientation at each pyramid resolution level. To extract the feature from each sliding window, we have divided it into cells at several pyramid level. The grid has 4N individual cells at N resolution level (i.e. N=0, 1, 2..). Histogram of gradient orientation of each pixel is calculated from these individual cells and is quantized into L bins. Each bin indicates a particular octant in the angular radian space. The concatenation of all feature vectors at each pyramid resolution level provides the final PHOG descriptor. In our implementation, we have limited the level (N) to 2 and we considered no of bins (L) to 8. As a result we obtained (1×8) + (4×8) + (16×8) = (8+32+128) = 168 dimensional feature vectors for individual sliding window position. For multi-oriented text this movement of sliding window should follow the polynomial equation fitted for the multi-oriented word image as mentioned in [18, 28]. The feature extracted from a sliding frame is shown in Fig. 10.

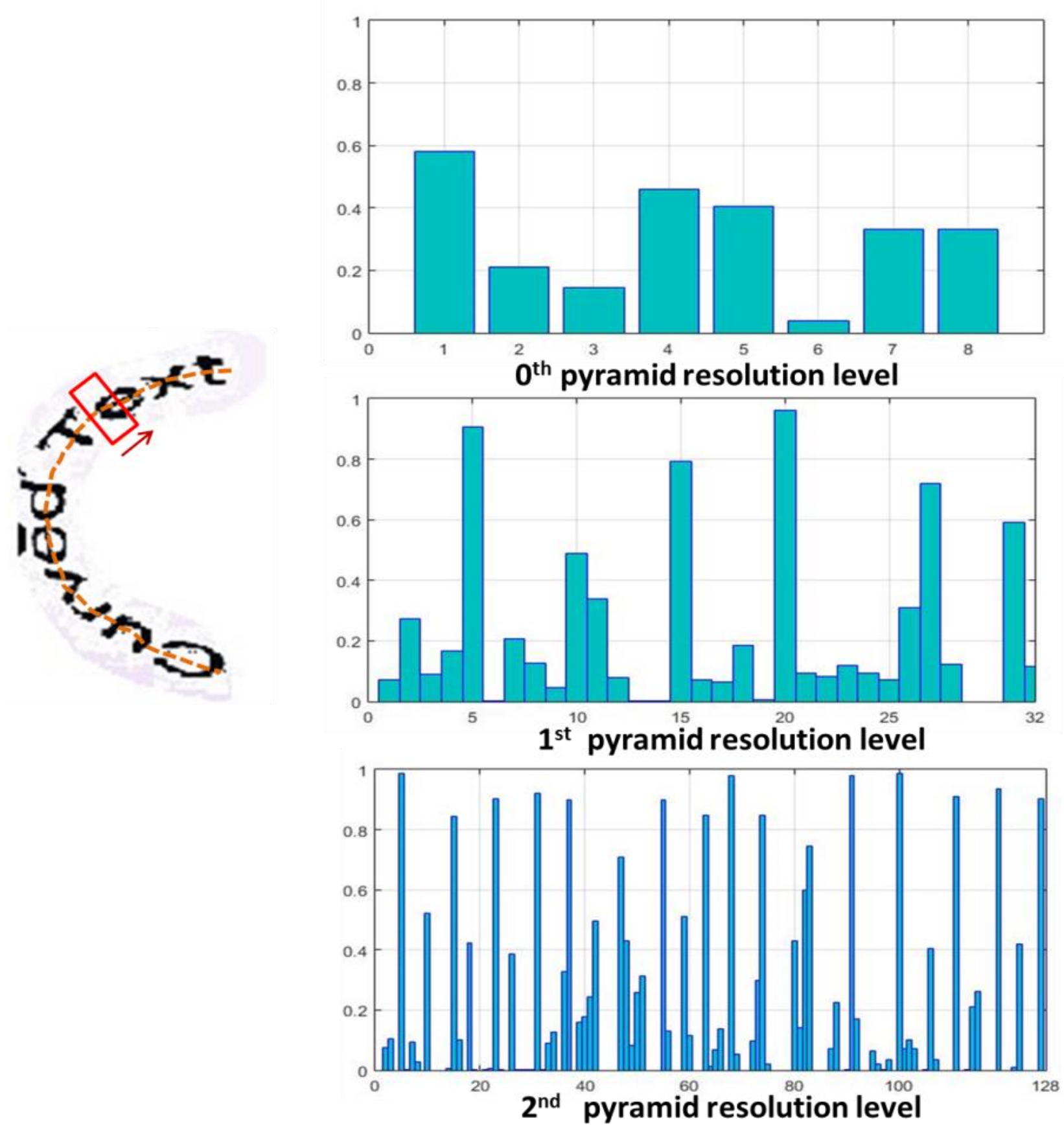

**Fig.10. (a) PHOG feature extraction from curved image following a polynomial equation. (b) PHOG feature at different pyramid levels for sliding window position marked in red.**

### 3.2.2. Hidden Markov Model-based Text Verification

HMM is a sequential classifier. An HMM may be defined by its initial state probabilities, state transition matrix and output probability matrix. Each state of model is associated with a separate Gaussian Mixture Model (GMM). For a classifier of *C* categories, we choose the model which best matches the observation from *C* HMMs:

$$\lambda_m = \{A_m, B_m, \pi_m\} \text{ .......... (5)}$$

where $m = 1 \ldots. C$, and $\sum_{m=1}^{C} \lambda_m = 1$. This means when a unknown sequence of unknown category is given, we calculate P $(\lambda_i \mid O)$ for each HMM $\lambda_m$ and select $\lambda_c^*$ , where $c^*$= arg $\max_m P(\lambda_m \mid O)$

$$P(\lambda_i|O) = \frac{P(O|\lambda_m)P(\lambda_m)}{P(O)} \text{ .......... (6)}$$

where, $P(O)$ is the density function for x irrespective of the category and is computed by:

$$P(O) = \sum_{m=1}^{C} P(O|\lambda_m)P(\lambda_m) \text{ .......... (7)}$$

The term $P(O|\lambda_m)$ is called the likelihood function for $O$ given, $\lambda_m . P(\lambda_m)$ is called the marginal or prior probability of $\lambda_m$.The standard solution, performed by the Viterbi algorithm, computes probability $P(O|\lambda_m)$of that sequence generated by $\lambda$ and finds the best likelihood of a class for a given feature vector sequence.

The HMM verification gives us a probability score for each text/non-text skeletons. The skeletons which give a probability score lesser than a threshold $T_v$, are removed from future processing. After conducting various tests over a significantly huge number of images varying over a large range of complex backgrounds, font sizes and orientation of text regions, we have set the value of $T_v$ as 0.44. Now, a situation may so arise, where more than one skeletal segment is found at the considered endpoint, all suitable for joining. In that case, the HMM probability score of each of the skeletal component, surrounding that particular junction point is obtained. From all the respective probability scores obtained, the skeletons whose scores are less than the set threshold value are eliminated permanently. Out of the remaining probability scores, the top 2 scores are chosen and the two skeletons corresponding to them are joined linearly and appended to the output-file image. The process discussed above is briefly described in Fig.11.

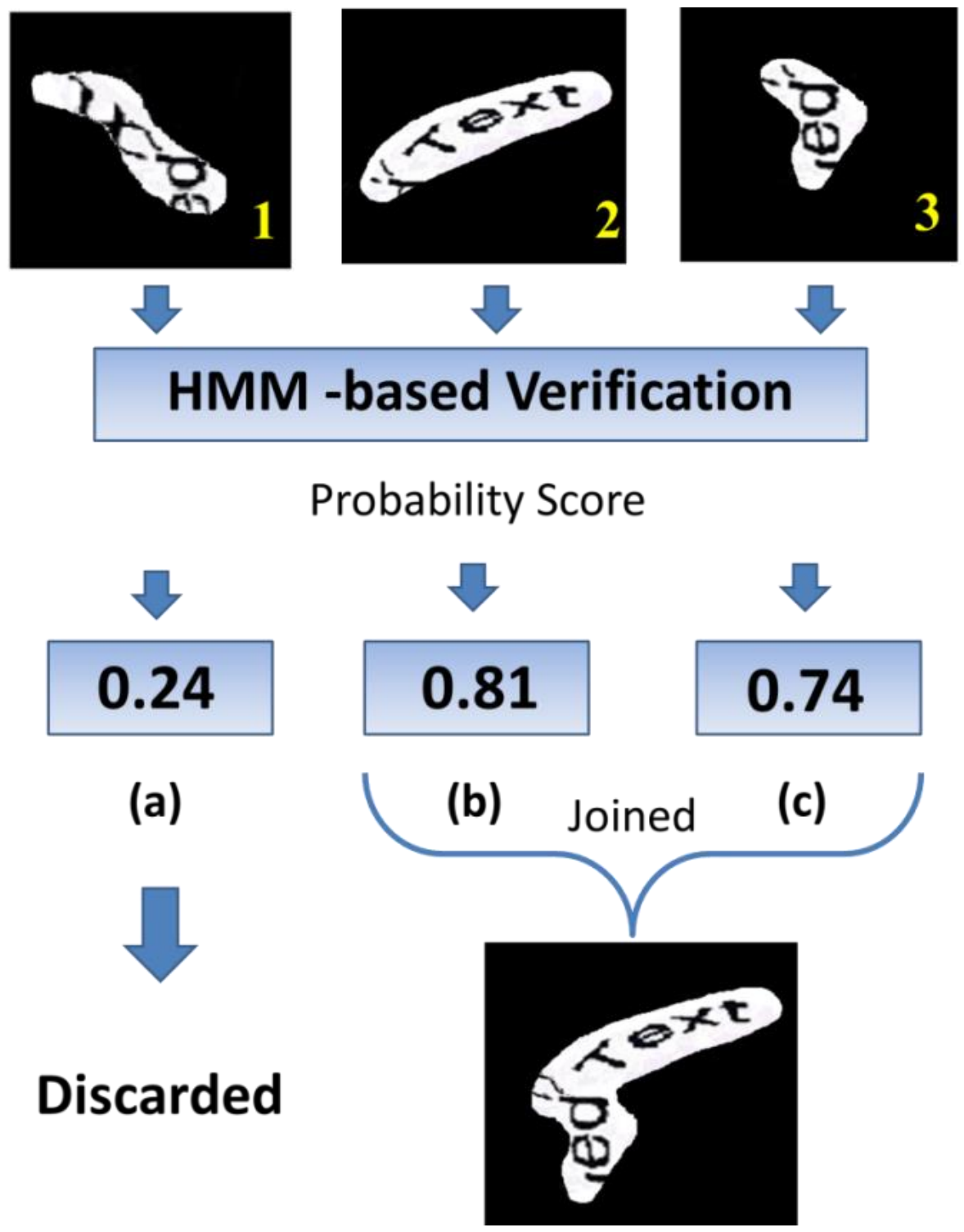

**Fig.11. Selection of the best possible combination using HMM probability score. The text-line images are taken from Fig.8.**

The above steps reduce the number of possible skeletons that might be encountered at the next junction point. Hence the number of skeletons to be checked again for the presence of text, at that junction point, is reduced. These above factors combined makes this proposed approach an efficient system of searching for candidate text regions via existing skeletons.

## 4. Experimental Results and Analysis

In this section we discuss in detail the performance analysis of our method on various datasets. We have also compared our method's performance with other existing methods to demonstrate the efficiency of the proposed algorithm. The following sub-sections deal with information regarding the various datasets that has been used for the experiments, qualitative analysis on those image sets that showcase our performances, quantitative analysis shown in tabulated form, a comparative study on the performances of the methods involved and lastly error analysis.

### 4.1. Description of Datasets

Our method focuses on locating the text region in a given image or a video frame and extracting the found text region, irrespective of the language of the script or its orientation. To depict these features we have chosen datasets that contain texts of various languages like Chinese, Japanese, Devanagari and Bengali, which shows the portability of the proposed framework in different scripts. Secondly, we have chosen datasets which contain images having the text area in any arbitrary orientation. Horizontal and non-horizontal texts to some extent are usually seen in most day to day images, but adding to that we have also included images of curved texts in our dataset which showcases the orientation-independent text-detection feature of our method. We name it as IITR-Text dataset. This dataset is made publicly available for further research in this direction [46].

We have performed the experiments on broadly three types of text datasets namely, horizontal texts, non-horizontal but linear texts and curvilinear texts. Initially we have performed the tests on a set of images gathered as frames from news-videos available on YouTube, and other scene texts from various sources which constitute our own dataset. Horizontal text images are also widely available and we have selected a number of images to contribute to our own data set. Non-horizontal and curved text datasets are not very readily available, which is why we had to gather images containing such text from YouTube videos, news videos, sport-videos, movie clips and from other sources. Qualitative analysis further down would showcase the performances of our algorithm on such images. The components of our in-house dataset are shown in the Table I. Some examples of different scripts are shown in Fig.12.

**Table I: Overview of IITR-Text detection dataset**

| Shape of Text | Scripts | | | |
|---|---|---|---|---|
| | English | Chinese | Devanagari | Bangla |
| Horizontal | 500 | 220 | 56 | 74 |
| Linear Non-Horizontal | 250 | 80 | 43 | 32 |
| Curved | 123 | 83 | 32 | 21 |

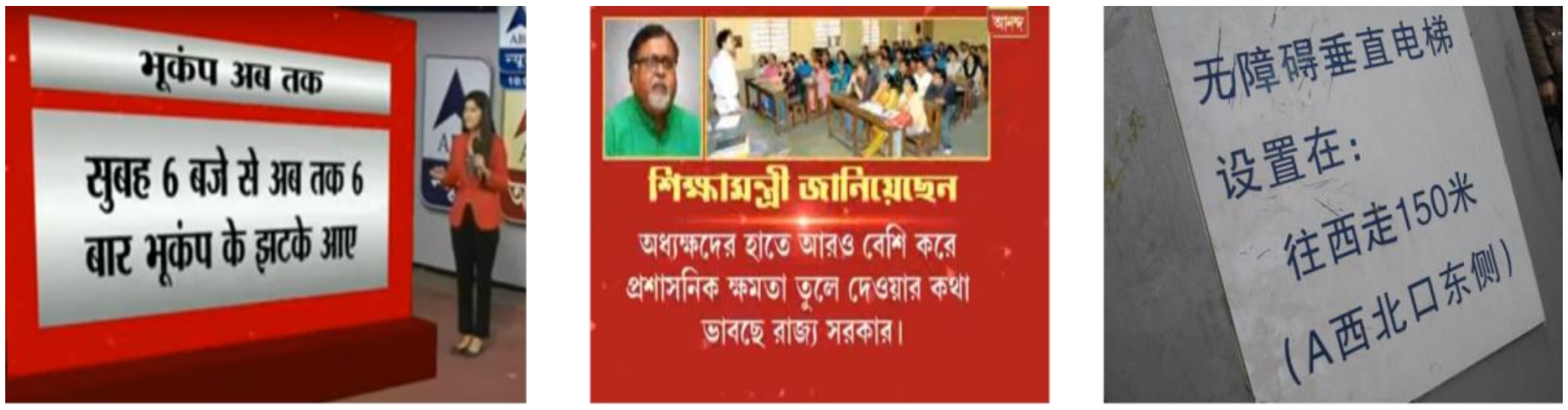

(a) (b) (c)

**Fig.12. Examples of non-Latin text image from IITR-Text dataset. (a) Devanagari (b) Bangla (c) Chinese.**

We have also used public datasets to have a common platform for comparison of efficiencies of various methods in this endeavor along with ours.The natural scene databases used were ICDAR 2013 [39] which has 462 images, SVT [41] which has 350 and MSRA-TD500 [42] which has 500 images. These datasets contain a large number of scene texts on which our algorithm has been tested and the results have been compared with other existing methods. The ICDAR 2013 scene-images include high resolution, complex backgrounds with mostly horizontal texts. The Street View Text (SVT) dataset focuses on street view images where complex backgrounds comprising of greenery and buildings can be seen. In this dataset, most of the text is in a horizontal direction. The MSRA-TD500 dataset covers a variety of scene text recorded in indoor and outdoor environments and the text is in multi-oriented directions. Along with these datasets we have also included a recent USTB-SV1K dataset [24, 51] for comparison purpose. Recently, this dataset has drawn attention of many researchers owing to its practical challenging and diverse range of images. This dataset contains 1000 (500 for training and 500 for testing) street view (patch) multi-oriented images from 6 USA cities. This dataset, USTBSV1K [24, 51] has a general open and challenging situation for natural (street view) scene text detection and recognition.

To test our algorithm on videos we have used the ICDAR 2015 [38] database which has 24 videos, 12–15s long, resulting in 10,800 frames, ICDAR-2013 videos database [39] which has 15 videos, 10–15s long, giving 7000 frames and YVT [40] videos which have 30 videos, 12–16s long, resulting in 13,500 frames. The ICDAR-2015 video dataset, which is basically an extension of ICDAR 2013 video database, has much more diversified text images. The ICDAR 2013 video dataset showcases a large variety of texts in terms of font-type, font-size, resolution alterations, complexity of backgrounds and difference in orientations. Graphics and scene-texts are provided by both ICDAR-2015 and ICDAR-2013 videos. The YVT dataset consists of high resolution videos carrying a large variety of scene text with multi-faceted backgrounds. The objective behind choosing the above diverse dataset is to demonstrate that the proposed algorithm works effectively in different situations. Some examples of these images are shown in Fig. 13.

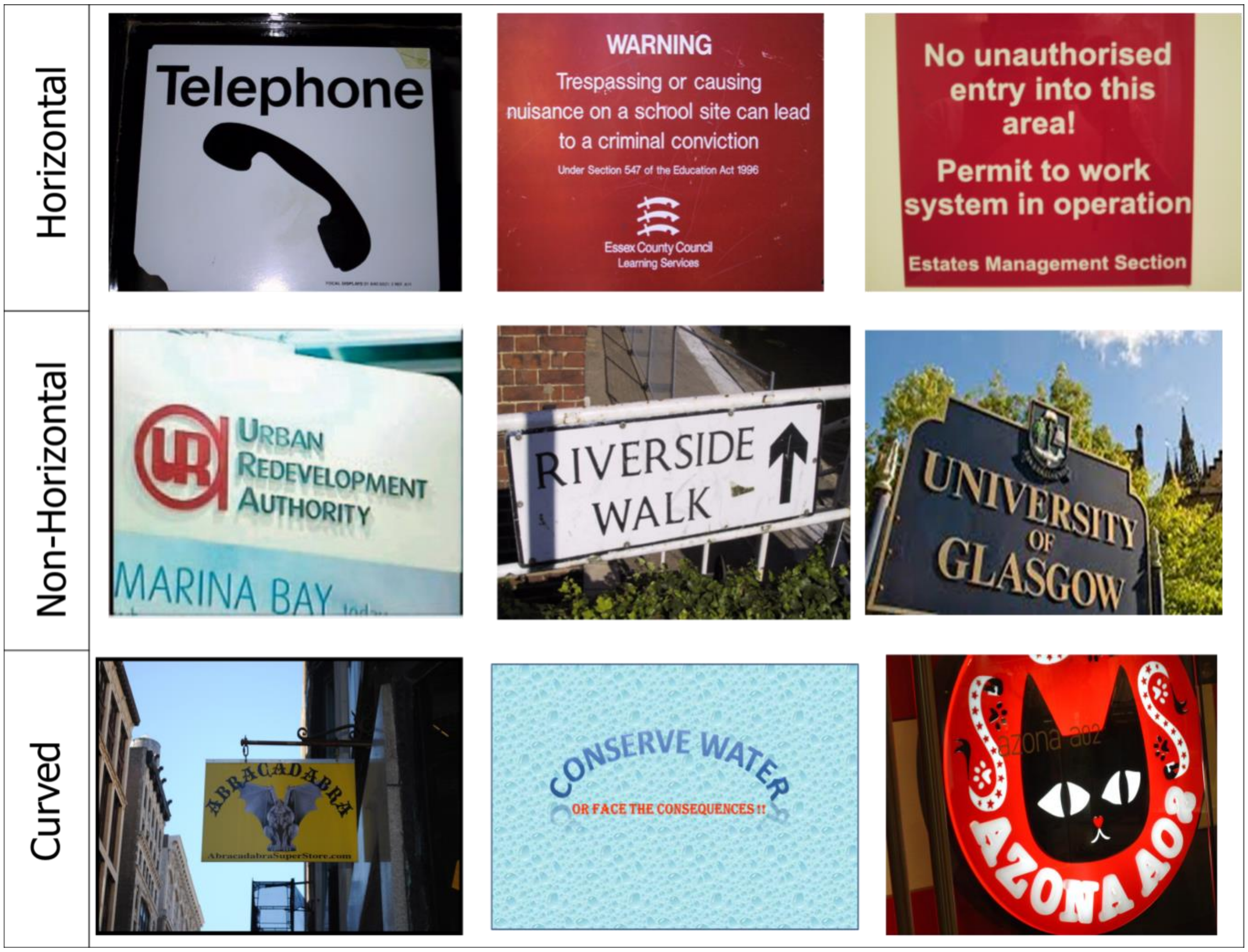

**Fig 13: Examples showing images from IITR-Text datasets (a) Horizontal text image (b) Linear Non-horizontal text image. (c) Curved text image.**

### 4.2. Evaluation protocol

To evaluate our performance, the following protocol has been chosen. Three classes are defined for each detected block: (i) 'Truly Detected Block (TDB)' – The truly detected text block (TDB) is a detected block that contains text partially or fully; (ii) 'Falsely Detected Block (FDB)' – The falsely detected text block (FDB) is a block that does not contain any text; and (iii) 'Missing Data Block (MDB)', a text block that misses more than 20% of the characters in a text line. Also 'Actual Text Blocks (ATB)' is counted for each image. The performance measures are defined as follows:

- Recall (R) = TDB/ATB
- Precision (P) = TDB (TDB + FDB)
- F-measure (F) = 2×P×R/(P + R)

To make a fair comparison with other existing approaches, we have followed the standard evaluation scheme given in the ICDAR 2013 robust reading competition [39]. For comparison, we have used the four existing methods described in (i) Liu et al. [9] which extracts six statistical features from four Sobel edge maps, (ii) Cai et al. [20], which performs Sobel edge detection in the YUV color space and applies two text area enhancement filters, (iii) Wong et al.[14], marked as the gradient-based method, that finds the maximum gradient difference to locate text regions and lastly (iv) Shivakumara et al. [1] which performs text detection as explained in that paper. The same parameter values are used for all of the experiments. ICDAR 1 and ICDAR 2 are the top two winners of ICDAR 2005 competitions. Along with these papers, a comparative study is performed with recent state of the art methods in publicly available datasets to justify the competency of our proposed method.

### 4.3. Experiment on IITR-Text Dataset

#### 4.3.1. Results on Horizontal text

A total of 850 images have been selected in our experiment. The English sub-data set contains 500 images (363 graphic-text images and 137 scene-text images), and the Chinese sub-data set contains 220 images (135 graphic-text and 85 scene text). Some examples in Fig.14 show the better performance of our method on horizontal text than the previous method mentioned in [1]. We have also selected 74 Bengali and 56 Devanagari scripted video frames to test the method on various linguistic scripts. Results are shown in Table II.

**Table II: Results on horizontal text detection on IITR-Text dataset**

| Method | English | | | Chinese | | |
|---|---|---|---|---|---|---|
| | R | P | F | R | P | F |
| Liu et al. [9] | 0.62 | 0.65 | 0.64 | 0.72 | 0.67 | 0.69 |
| Cai et al. [20] | 0.53 | 0.42 | 0.46 | 0.58 | 0.39 | 0.47 |
| Wong et al. [14] | 0.62 | 0.81 | 0.70 | 0.62 | 0.78 | 0.69 |
| Shivakumara et. al.[1] | 0.77 | 0.82 | 0.80 | 0.75 | 0.79 | 0.77 |
| Neumann et al. [37] | 0.79 | **0.89** | 0.84 | 0.78 | **0.87** | **0.83** |
| **Proposed** | **0.81** | 0.88 | **0.85** | **0.79** | 0.85 | 0.82 |
| **Method** | **Bangla** | | | **Devanagari** | | |
| | R | P | F | R | P | F |
| Liu et al. [9] | 0.59 | 0.62 | 0.61 | 0.52 | 0.65 | 0.58 |
| Cai et al. [20] | 0.49 | 0.42 | 0.42 | 0.42 | 0.35 | 0.38 |
| Wong et al. [14] | 0.65 | 0.83 | 0.72 | 0.59 | 0.72 | 0.64 |
| Shivakumara et al.[1] | 0.75 | 0.83 | 0.79 | 0.73 | 0.81 | 0.77 |
| Neumann et al.[37] | 0.78 | 0.86 | 0.82 | 0.77 | **0.85** | 0.81 |
| **Proposed** | **0.82** | **0.86** | **0.84** | **0.80** | **0.85** | **0.82** |

### 4.3.2. Results on Linear non-horizontal text

Here, we have selected a total of 405 images from our own dataset. The English set contains 250 images (175 for scene-text images and 75 graphic-text images), Chinese sub-data set contains 80 images (47 for graphics text and 33 for scene text). The images for Bengali (32 images) and Devanagari (43 images) scripts have been selected from you-tube and other video sources. As we see from Fig.15, our algorithm successfully depicts the text in the oblique direction quite effectively. The results are compared in Table III, along with the results on curve text detection as well which is described in the next section.

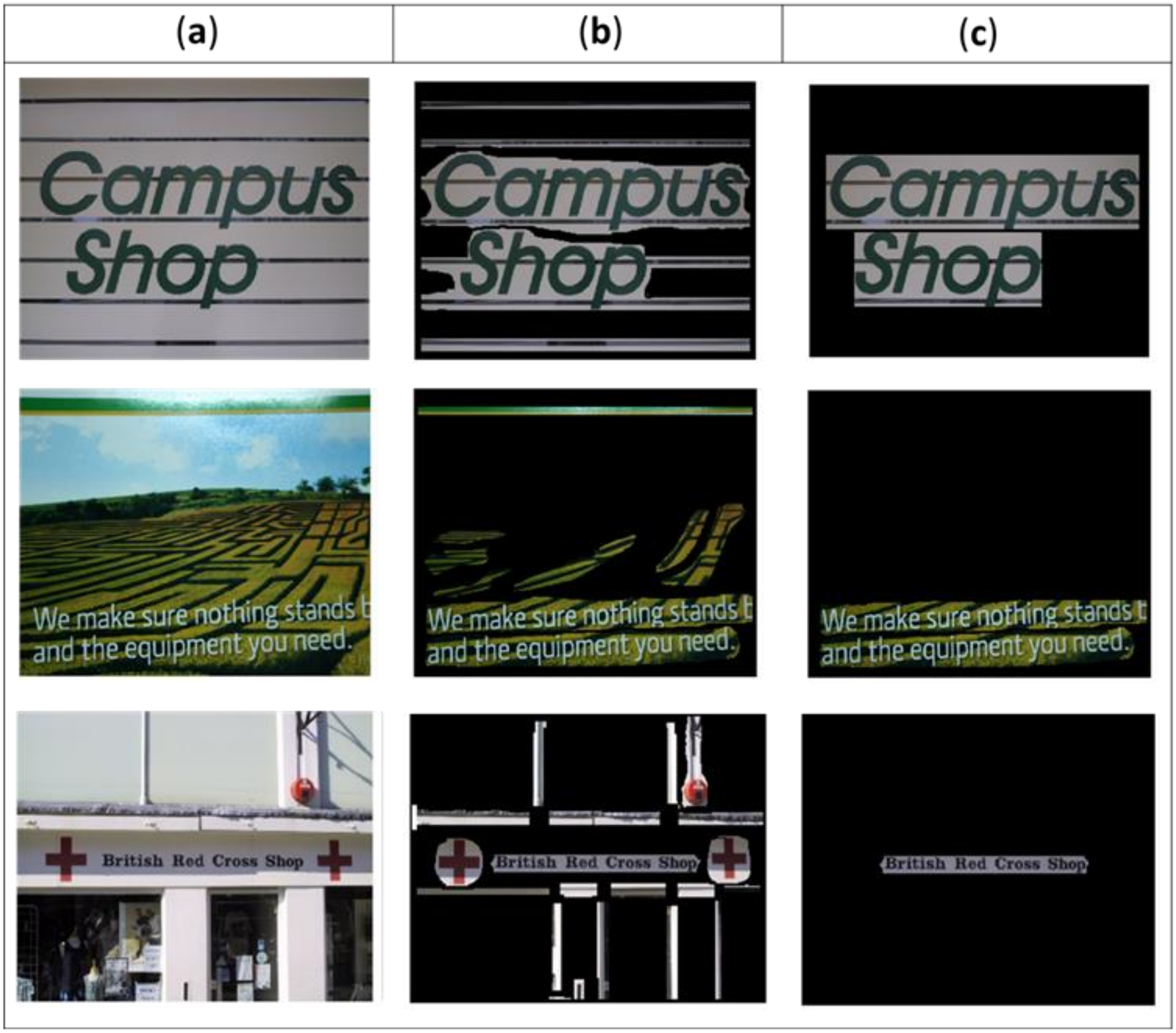

**Fig 14: Qualitative results on horizontal text. (a) Original image (b) Shivakumara et al. [1] (c) Our Method.**

### 4.3.3. Results on curved text

In this case, not many curved texts are available to test the method upon, hence custom datasets acquired from various videos and other scene texts have been used for testing. The English dataset contains 123 images, Chinese dataset contains 83 images, Bengali set contains 21 images and Devanagari contains 32 images. As is evident from Fig. 15, other methods cannot detect the curved areas properly which our method can, with certain accuracy. The HMM verification is responsible for a higher accuracy rate here. The corresponding results are compared in Table III.

**Table III: Result on linear non-horizontal and curved text detection on IITR-text dataset**

| **Linear Non-horizontal Text** | | | | | | |
|---|---|---|---|---|---|---|
| **Method** | English | | | Chinese | | |
| | R | P | F | R | P | F |
| Shivakumara et al.[1] | 0.72 | 0.77 | 0.74 | 0.69 | 0.73 | 0.71 |
| Neumann et al. [37] | 0.75 | 0.83 | 0.79 | 0.72 | 0.82 | 0.77 |
| **Proposed** | **0.79** | **0.85** | **0.82** | **0.75** | **0.83** | **0.79** |
| Method | Bangla | | | Devanagari | | |
| | R | P | F | R | P | F |
| Shivakumara et al.[1] | 0.74 | 0.78 | 0.76 | 0.73 | 0.79 | 0.76 |
| Neumann et al. [37] | 0.70 | 0.76 | 0.73 | 0.72 | 0.77 | 0.75 |
| **Proposed** | **0.80** | **0.82** | **0.81** | **0.79** | **0.83** | **0.81** |
| **Curved Text** | | | | | | |
| **Method** | English | | | Chinese | | |
| | R | P | F | R | P | F |
| Shivakumara et al. [1] | 0.52 | 0.59 | 0.55 | 0.53 | 0.61 | 0.57 |
| Neumann et al.[37] | 0.59 | 0.63 | 0.61 | 0.59 | 0.64 | 0.62 |
| **Proposed** | **0.74** | **0.81** | **0.77** | **0.71** | **0.77** | **0.74** |

### 4.3.4. Improvement of Text Detection Result by HMM-Verification

HMM verification of detected text areas increases the accuracy or the detection rate. It compares the given probable text segment with the text and non-text models and thus verifies the region as a text part or non-text part. HMM verification scheme has been found highly essential for the curve text detection, as, in this case we are not assuming that the text appears in straight lines only unlike the method in Shivakumara et al.[1]. In such a case, without HMM based verification the method would yield unsatisfactory results.

Here, a total of 4,326 image patches were considered to train the HMM model for text verification. A total of 2,345 image patches are used from text regions whereas, rest 1981 image patches are considered from non-text region. During HMM-based verification, we have noted that state number 6 and Gaussian number 32 provide best results in the validation dataset.

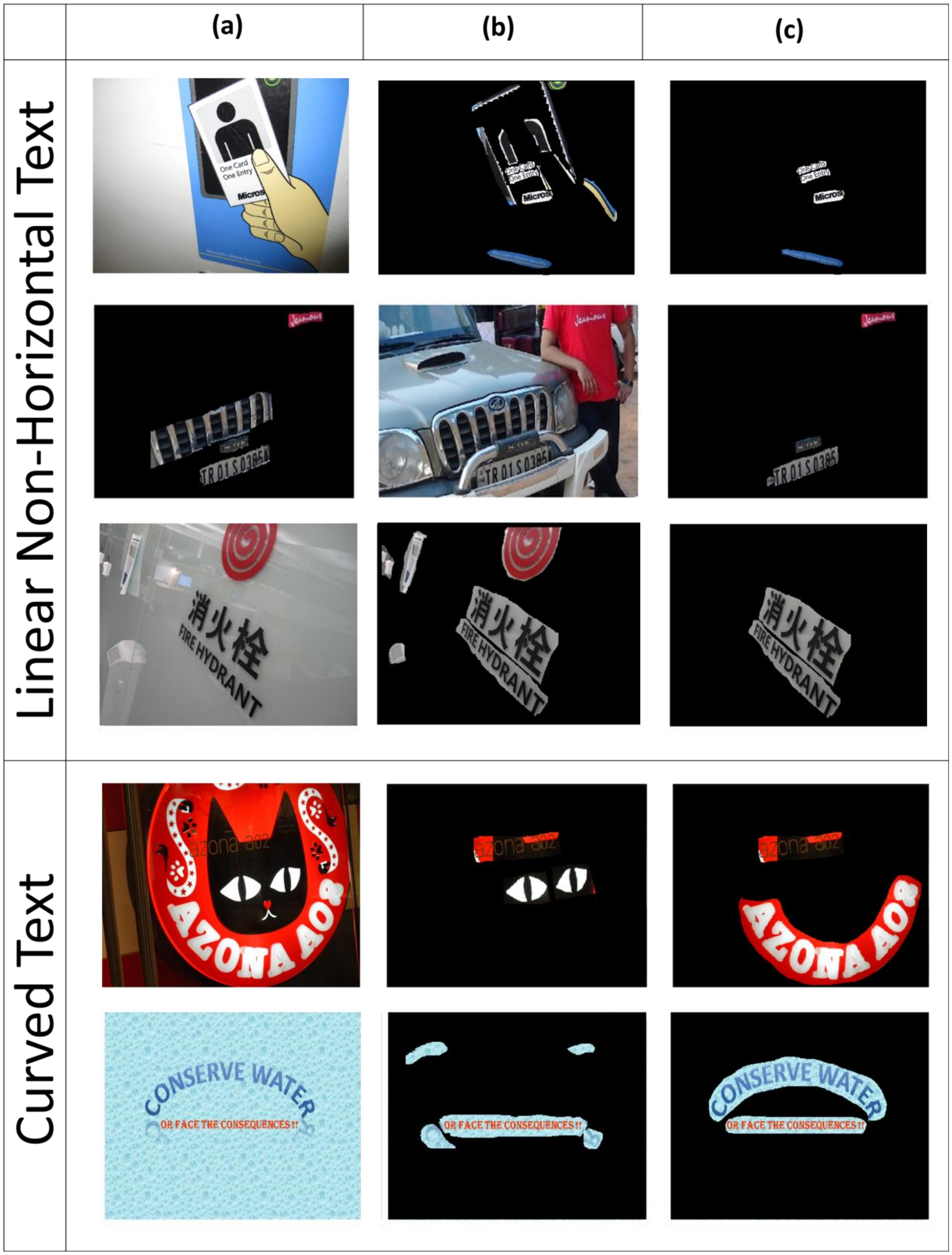

**Fig 15: Qualitative results on linear non-horizontal text and Curved Text (a) Original image (b) Shivakumara et al. [1]. (c) Our system.**

Also in cases of horizontal or linear-non-horizontal text detection, without the use of HMM-based verification step, there is a chance of including false positives. Such false positives might include parts of logo or other complex linear features as shown in Fig 13(b) and Fig 14(b). But on using HMM-based verification chances of including such false positives are minimized significantly.

As we can see from the Table IV, the recall value in the case of text detection without HMM-based verification is significantly higher than the one with HMM verification (the method that we have applied), but its precision values are quite low. This can be understood simply if we note the fact that without HMM verification we are considering all the encountered skeletons which are produced after skeletonization of candidate text regions. Along with all possible text fields the candidate text regions contain a huge number of non-text areas as well. This explains why the precision value comes out to be so low. But when we use HMM verification, the chance of including non-text areas reduce significantly.

**Table IV: Comparison of Text/Non-text classification efficiency using HMM verification on IITR-text dataset (English only)**

| Our system | Horizontal Text | | | Linear non-horizontal Text | | | Curved Text | | |
|---|---|---|---|---|---|---|---|---|---|
| | R | P | F | R | P | F | R | P | F |
| With HMM | 0.81 | 0.88 | 0.85 | 0.79 | 0.85 | 0.82 | 0.74 | 0.81 | 0.77 |
| Without HMM | 0.92 | 0.73 | 0.81 | 0.87 | 0.67 | 0.75 | 0.85 | 0.54 | 0.66 |

A confusion matrix has been shown (see Fig.16) that denotes the text/non-text classification results of our algorithm used. Note that, in spite of a few anomalies, our text detection rates are quite high.

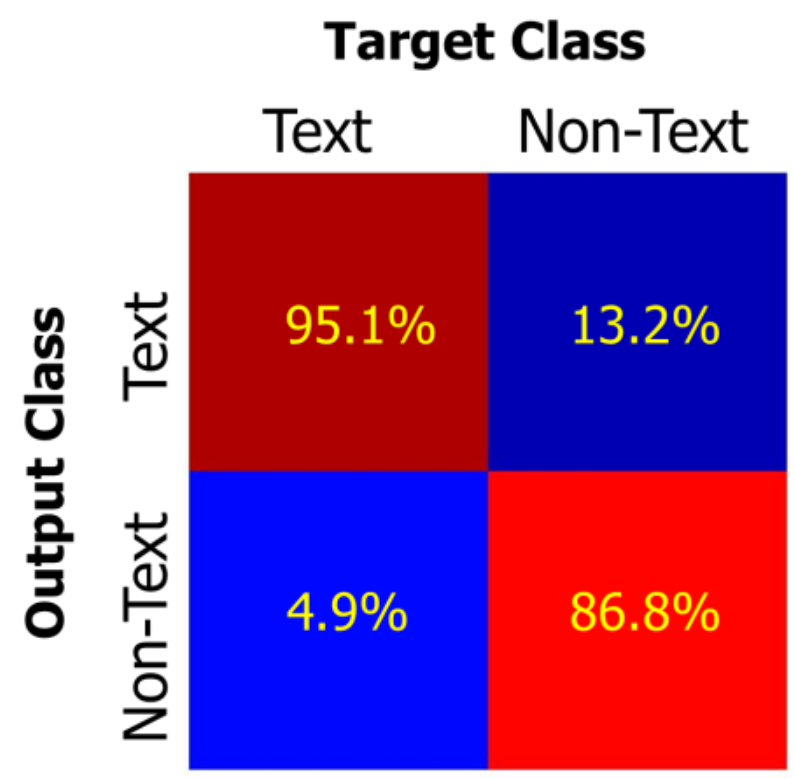

**Fig 16: Confusion Matrix between Text and Non-Text classification using HMM**

In this section we have taken mainly three methods, namely the ones described in (1) Shivakumara et. al. [1] which uses skeletonization to detect probable text regions but does not use HMM based verification techniques, (2) Neumann et. al. [37] which uses an end-to-end text localization and recognition method, and (3) our proposed algorithm. Our algorithm has been broken into 2 processes. One that uses HMM-based verification and another one that doesn't. Fig.17 shows their comparative performances, the respective F-measure values of their performance on our own IITR English text dataset for horizontal, linear non-horizontal and curved texts.

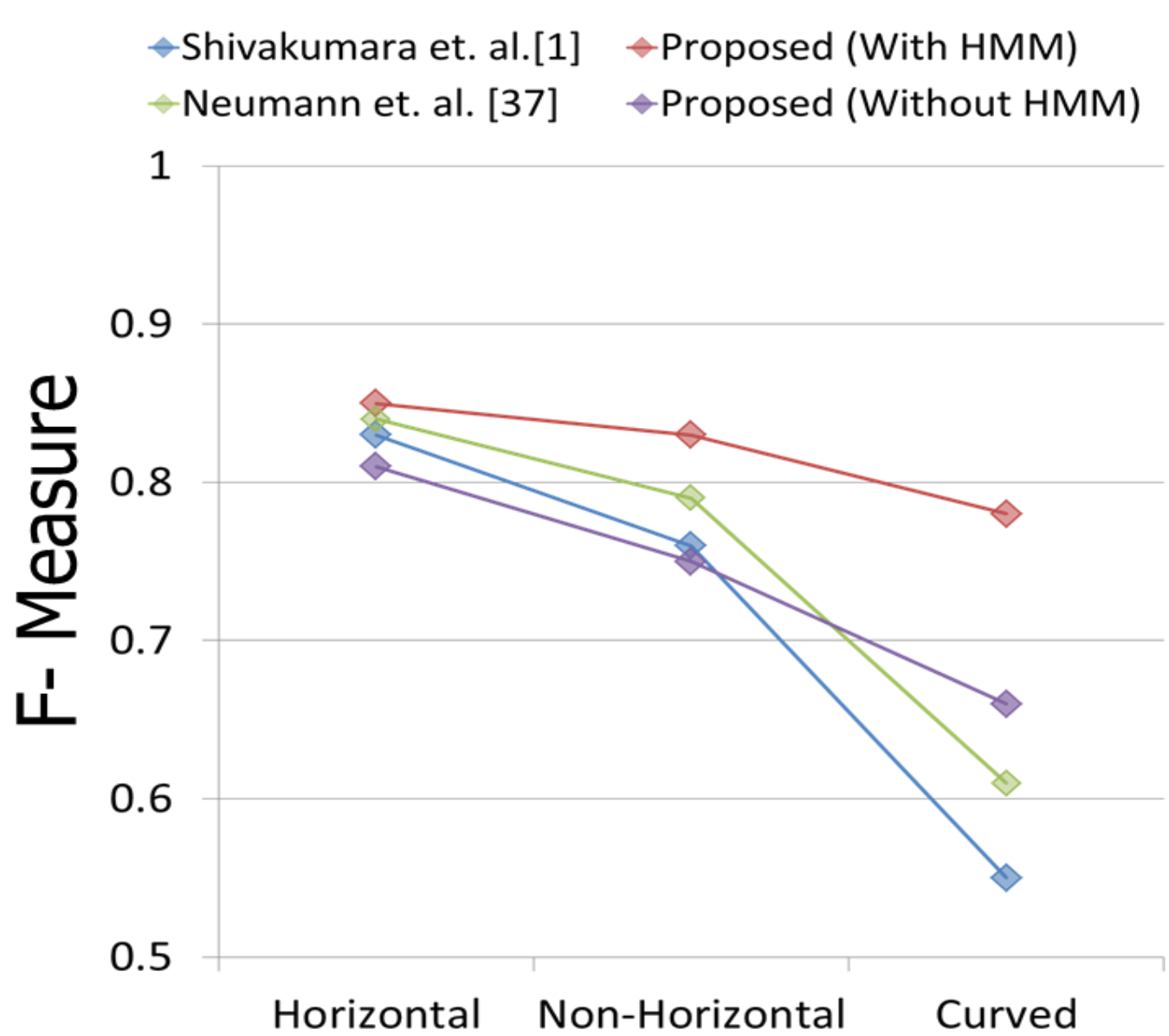

**Fig 17: Graphical Representation of performances of 4 methods in Shivakumara et. al. [1], Neumann et al. [37], and ours, with HMM verification and without it, on IITR-Text dataset.**

### 4.3.5 Comparison between LoG and Laplacian filtering

We have compared the performance of our Laplacian of Gaussian (LoG) filtering with Laplacian filtering alone. In images with texture or contour distortions, or in low contrast images with complex components, applying only Laplacian filter generates a lot of noise. Filtering such noise in the subsequent steps also removes some text information which is necessary for recovering the text region boundary. It is due to the fact that significant edge information could not be retrieved from the noisy or low contrast images. The LoG filter in the Fourier domain helps in retaining most of the text pixels. As mentioned earlier, if some background noises are included in that filtering step, they are gradually eliminated in the subsequent processes. For instance, isolated skeletal segments of those tiny patches, resulting from such noise, will be small enough to be neglected in further processing. Moreover, verification step in later stage will eliminate such noise-induced false positives.

Qualitative results obtained using these two different filtering techniques are shown in Fig. 18. Note that, the boundaries of the images in columns of Fig.18(b) are not well preserved with the Laplacian filtering. But with LoG filtering, the boundaries of the text regions (see images in columns of Fig.18(c)) obtained are maintained and are more acceptable. Thus LoG filter has been used for better performance in text detection. On using LoG filter instead of Laplacian filter, we have achieved an increment in the F-measure, by an amount of 0.09, 0.12 and 0.15 on the horizontal, linear non-horizontal and curved text data sets (IIT-R) respectively.

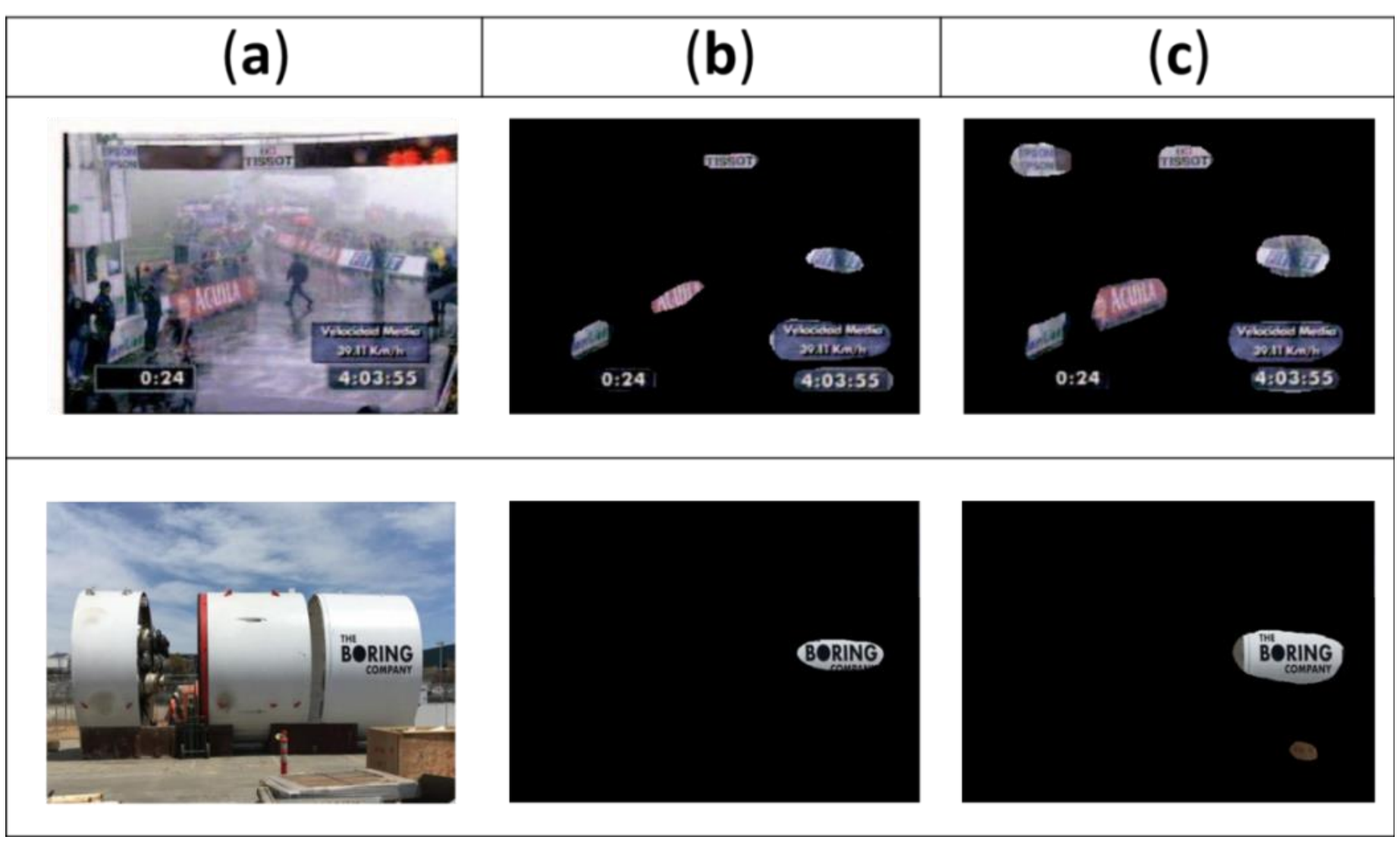

**Fig. 18: Comparison between two types of filtering used. (a) Original Image. (b) Laplacian Filtered (c) LoG filtered.**

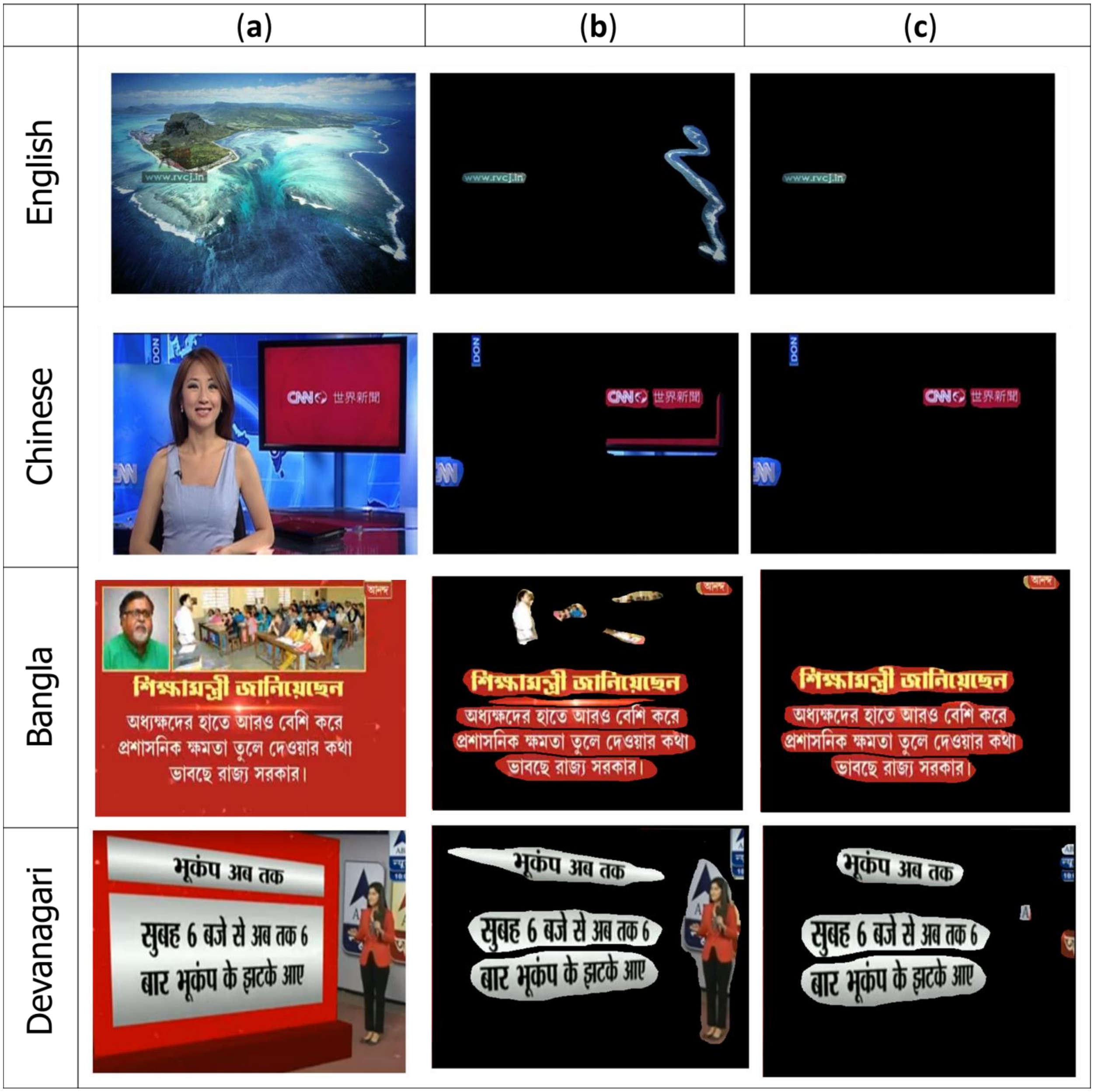

**Fig 19: Qualitative analysis: (a) Original image (b) Shivakumara et al. [1] (c) Our System**

### 4.3.6. Experiment on Multiple Scripts

Along with the experiments on English texted datasets we have also used our algorithm on various other non-Latin scripts, for instance Chinese, Japanese and Devanagari scripts. Fig.19 shows that our algorithm works equally well on non-Latin texts as it does on English ones.

### 4.4. Experiment on Public Datasets

To compare the efficiency of our method with the other existing methods of text detection we have performed experiments on a few public datasets. This provides us with a common platform for the other methods involved to showcase their effectiveness as well. The results of these methods that have been used for comparison purposes are tabulated below, in terms of their precision-recall rating. From the information given below we can understand that our method provides a better accuracy than the methods we have compared it with. This is due to the fact that our algorithm uses Hidden Markov Model for text verification.

#### 4.4.1 Performance on Video datasets

We have tested our algorithm on a few other datasets as well. This time we have compared our method's performance with a few more algorithms of different other papers, as detailed earlier. The highest performance results have been marked in bold. It is evident that our method scores higher than the methods mentioned above. This is because our method not only relies on contrast details but also has a dedicated focus on the orientation and direction of the text region present. Skeletonization and proper curve fitting ensures that the text region is correctly mapped and HMM verification adds a certain accuracy level to our results as is evident from the table (Table V) below.

**Table V: Comparative study on public video datasets**

| Methods | Datasets | | | | | | | | |
|---|---|---|---|---|---|---|---|---|---|
| | ICDAR 2013 | | | YVT | | | ICDAR 2015 | | |
| | R | P | F | R | P | F | R | P | F |
| Epshtein et al. [5] | 37.05 | 32.7 | 34.91 | 41.35 | 36.9 | 38.99 | 35.78 | 34.5 | 35.12 |
| Liu et al. [9] | 33.1 | 28.6 | 30.68 | 38.6 | 34.1 | 36.2 | 33.17 | 31.8 | 32.47 |
| Shivakumara et al. [21] | 47.3 | 44.92 | 46.07 | 59.8 | 57.4 | 58.57 | 55.7 | 52.3 | 53.9 |
| Zhao et al. [30] | 50.7 | 48.99 | 49.83 | 52.6 | 51.4 | 51.99 | 48.3 | 47.7 | 47.99 |
| Mi et al. [32] | 42.7 | 36.4 | 32.29 | 49.6 | 38.3 | 43.22 | 37.9 | 35.2 | 36.5 |
| Mosley et al.[57] | 49.0 | 50.0 | 49.0 | 79.0 | 72.0 | 75.0 | - | - | - |
| Wu et al. [56] | 68.0 | 63.0 | 65.0 | 73.0 | 81.0 | 77.0 | - | - | - |
| Roy et al. [58] | 57.27 | 64.3 | 60.58 | - | - | - | - | - | - |
| Wu et al.[59] | 75.0 | 51.0 | 61.0 | - | - | - | 72.0 | 48.0 | 58.0 |
| Shivakumara et al. [29] | **79.0** | 65.0 | 71.0 | 76.0 | 79.0 | 76.0 | - | - | - |
| **Proposed** | 74.2 | **87.4** | **80.26** | **79.2** | **85.7** | **82.3** | **80.6** | **84.5** | **82.5** |

### 4.4.2. Performance on Scene-Image datasets

In the following table (Table VI) we show the Precision Recall and F-measure ratings of our performances on horizontal text images of the ICDAR 2003 dataset. As we can observe from the table above, that although Shivakumara et al. [1] shows a good accuracy, based on its precision-recall values, but our method has better accuracy. This is again owing to the involvement of HMM verification technique that we have implemented into our algorithm.

Similar results are also inferred from the observation table (Table VII) below, where our algorithm along with the other previous methods are tested on the MSRA-TD500 public datasets, along with a few other datasets. USTB-SV1K being a difficult dataset, we did not obtain high performance values in Table VIII, as observed in the results of other such datasets. However, amongst the methods that have been compared the proposed method shows the highest F-measure value.

**Table VI: Performance of horizontal text detection on ICDAR 2003 Dataset**

| Methods | R | P | F |
|---|---|---|---|
| Liu et al. [9] | 0.48 | 0.65 | 0.55 |
| Cai et al. [20] | 0.60 | 0.39 | 0.48 |
| Wong et al. [14] | 0.56 | 0.79 | 0.66 |
| Shivakumara et al.[1] | 0.85 | 0.79 | 0.82 |
| ICDAR 1 | 0.63 | 0.63 | 0.60 |
| ICDAR 2 | 0.58 | 0.58 | 0.56 |
| **Proposed** | **0.86** | **0.83** | **0.85** |

**Table VII: Comparative study of Horizontal text detection on public scene-text datasets.**

| Methods | Datasets | | | | | | | | |
|---|---|---|---|---|---|---|---|---|---|
| | MSRA TD500 | | | ICDAR 2013 | | | SVT | | |
| | R | P | F | R | P | F | R | P | F |
| Epshtein et al. [5] | 61.3 | 60.2 | 60.74 | 51.7 | 47.88 | 49.35 | 55.8 | 53.46 | 54.6 |
| Liu et al. [9] | 59.2 | 57.6 | 58.38 | 53.63 | 48.3 | 50.82 | 56.71 | 54.9 | 55.79 |
| Shivakumara et al. [21] | 71.3 | 69.2 | 70.23 | 55.9 | 52.0 | 53.87 | 66.8 | 64.9 | 65.83 |
| Zhao et al. [30] | 70.8 | 66.9 | 68.79 | 63.7 | 62.1 | 62.88 | 58.3 | 57.2 | 57.74 |
| Mi et al. [32] | 63.1 | 61.4 | 62.23 | 54.2 | 51.7 | 52.92 | 53.8 | 52.1 | 52.3 |
| Neumann et al. [37] | - | - | - | 72.4 | 82.3 | 77.03 | - | - | - |
| Liu et al[36] | - | - | - | 70.4 | **89.1** | 79.06 | - | - | - |
| Zhang et al. [34] | - | - | - | 74.5 | 88.6 | 80.94 | - | - | - |
| Text_Flow[33] | | | | 75.89 | 85.15 | 80.25 | | | |
| Dey et al.[54] | **85.0** | 52.0 | 65.0 | **87.0** | 67.0 | 75.0 | **68.0** | 55.0 | 61.0 |
| Zhang et al.[55] | 67.0 | 83.0 | 74.0 | 78.0 | 88.0 | 83.0 | - | - | - |
| Yin et al. [51] | 63.0 | 81.0 | 71.0 | 65.11 | 83.98 | 73.35 | - | - | - |
| Cho et al. [53] | - | - | - | 78.45 | 86.26 | 79.02 | - | - | - |
| Tian et al.[25] | - | - | - | 83.98 | 83.69 | **83.84** | - | - | - |
| **Proposed** | 79.2 | **85.7** | **82.3** | 74.2 | 87.4 | 80.26 | 67.68 | **81.65** | **74.01** |

**Table VIII: Performance of text detection on USTB-SV1K Database**

| Methods | R | P | F |
|---|---|---|---|
| Yin et al.[51] | 45.41 | 49.85 | 47.53 |
| Yin et al. [24] | 45.18 | 45.00 | 45.09 |
| Tian et al.[25] | **48.75** | 53.82 | 51.15 |
| Proposed | 47.32 | **57.88** | **52.14** |

### 4.4.3. Qualitative Results on Public Datasets

Experiments on a few images from the public datasets mentioned above have been shown below in Fig. 20, which depicts the operation of our method along with Shivakumara et al. [1]. As we can observe, the performance results on these public data sets are in coherence with the results obtained earlier from our own dataset. Looking at the overall scenario, it can be observed from the results that our proposed method performs fairly well in detecting horizontally oriented texts from images as compared to other methods, and very efficiently in detecting linear but non-horizontally oriented texts or texts oriented in curvilinear fashion from images.

### 4.5. Comparative study with other classifiers

In this paper, we have considered HMM for text verification purpose because of its ability to handle the sequential dependency. We have performed a comparative study of text verification performance with other existing classifiers [17], like Support Vector Machine (SVM), AdaBoost, Random Forest, and Conditional Random Field (CRF). The performance of various classifiers is evaluated in our IITR-text dataset which is tabulated in Table IX. Note that the average precision value is found to be highest in case of HMM. The improvement using HMM is mainly due to its better modeling of text information in a word.

**Table IX: Performance comparison using different classifiers on IITR-text dataset**

| Classifiers | Horizontal Text | | | Linear Non-Horizontal Text | | | Curved Text | | |
|---|---|---|---|---|---|---|---|---|---|
| | R | P | F | R | P | F | R | P | F |
| SVM | 0.79 | 0.80 | 0.79 | 0.76 | 0.77 | 0.76 | 0.69 | 0.71 | 0.70 |
| AdaBoost | 0.80 | 0.78 | 0.79 | 0.75 | 0.79 | 0.77 | 0.72 | 0.70 | 0.71 |
| Random Forest | **0.83** | 0.82 | 0.82 | **0.82** | 0.78 | 0.79 | **0.75** | 0.76 | 0.76 |
| CRF[11] | 0.80 | 0.85 | 0.82 | 0.79 | 0.81 | 0.80 | 0.71 | 0.78 | 0.74 |
| **Proposed** | 0.81 | **0.88** | **0.85** | 0.79 | **0.85** | **0.82** | 0.74 | **0.81** | **0.77** |

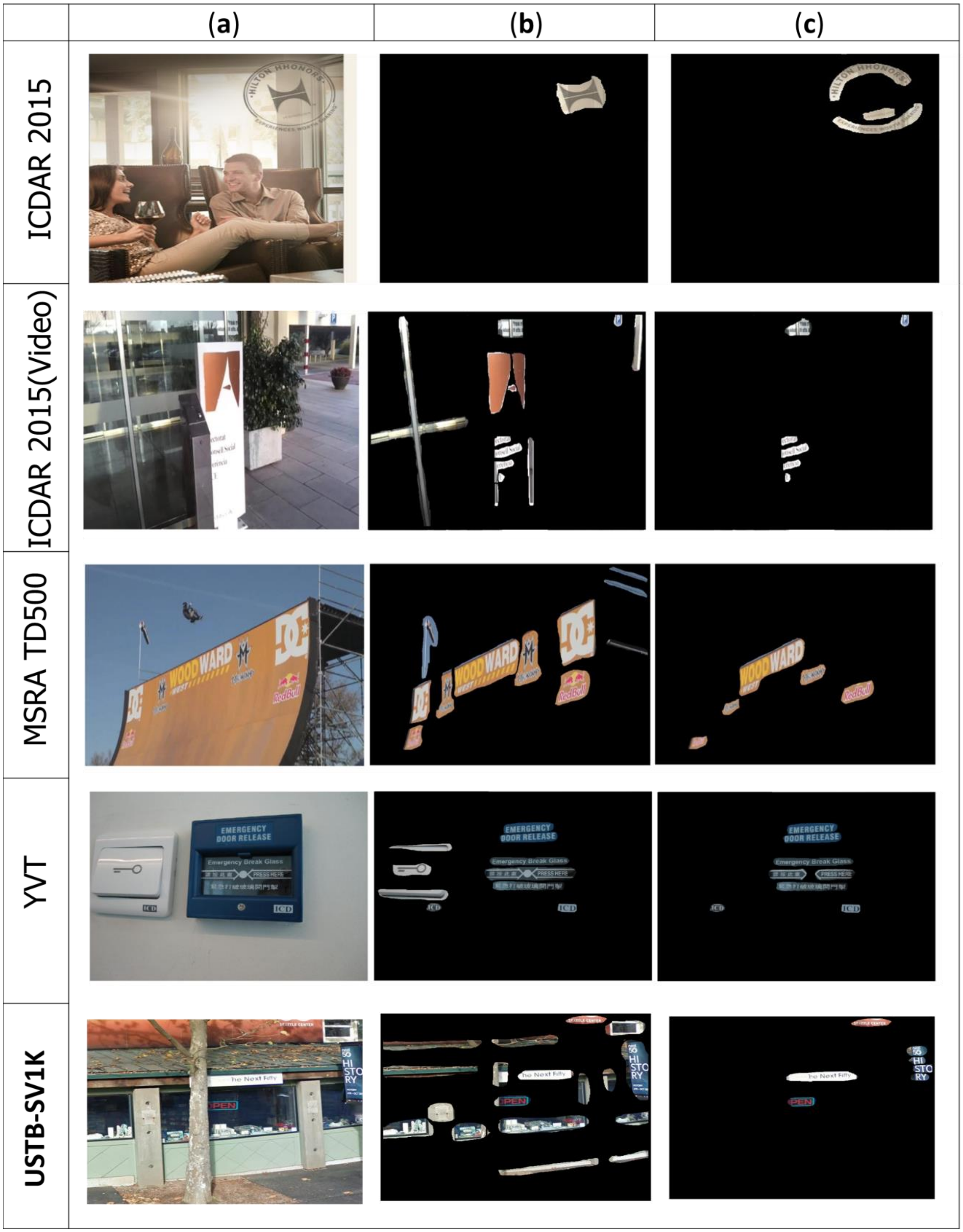

**Fig 20: Curved Text detection results on ICDAR 2015 dataset (a) Original image (b) Shivakumara et al. [1] (c) Proposed system**.

### 4.6. Parameter Evaluation

A comprehensive study is performed to find the optimum value of parameters and thresholds used in our experiments. During text identification stage, the parameter sliding window length in the Maximum Difference map, (mentioned in Section 3.1.1) measured in pixel units. From experimental study the lower limit of threshold was set to 7 pixels. The comparisons are shown in Fig 21(a). Secondly, a threshold value was considered for elimination of unnecessary skeletal fragments as discussed in Section 3.1.2. A skeletal segment of length below that certain value would be considered as noise. Experimentally it was found that a length of 15 pixels (See Fig. 21(b) removes the irrelevant small skeletal remnants in our framework. In text verification purposes using HMM, continuous density HMMs with diagonal covariance matrices of GMMs are used in each state. Our text detection performance in IITR-text dataset with different Gaussian numbers is shown in Fig.21(c). It was observed that 32 Gaussian Mixtures provided best results in our text detection framework. A comparative study with varying state numbers is shown in Fig. 21(d). The best state number was found to be 6.

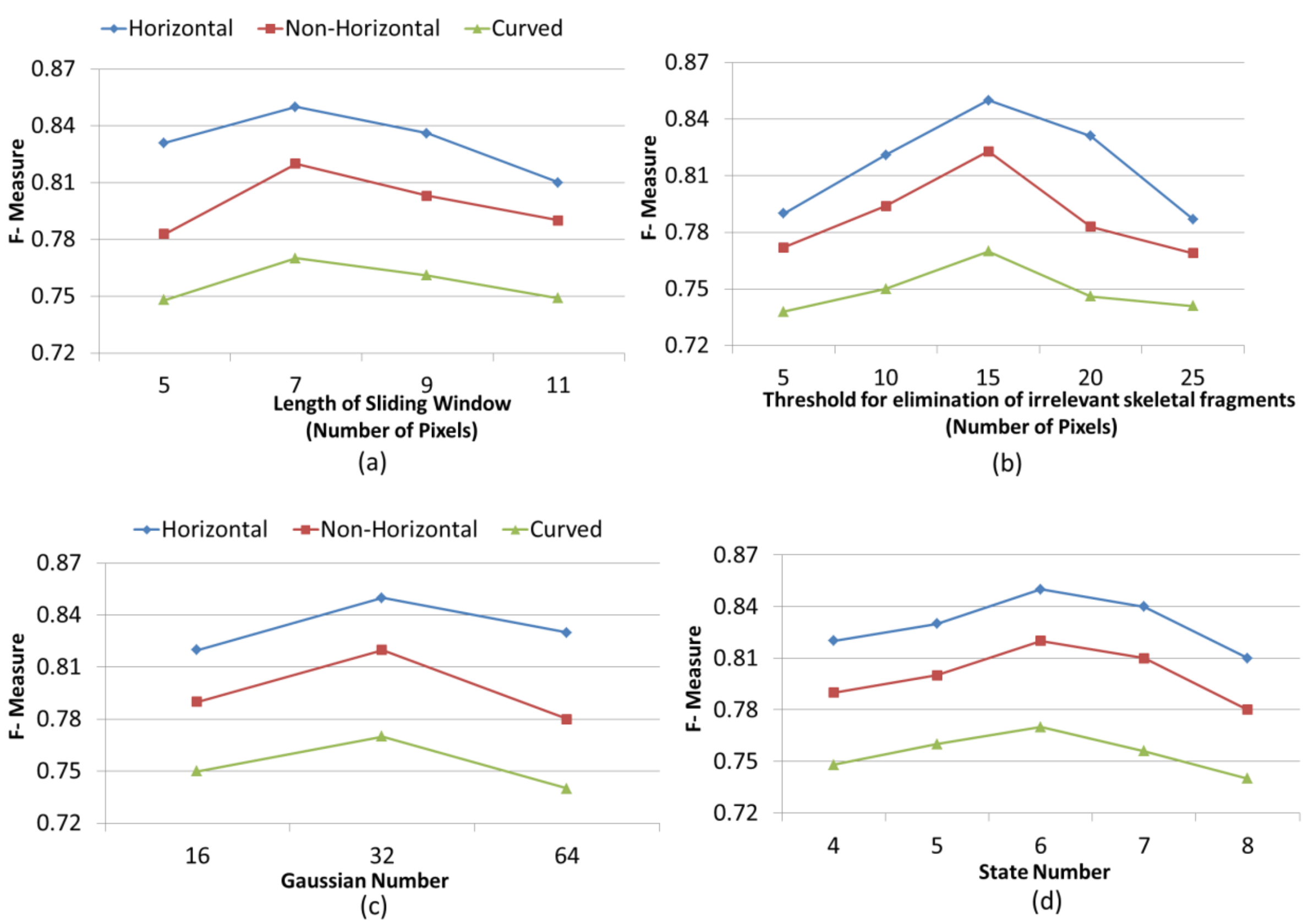

**Fig.21. Parameter evaluation based on our IITR text dataset (a) Length of Sliding Window (b) Threshold for elimination of irrelevant skeletal fragments (c) Gaussian number (d) State number.**

### 4.7. Experiment on Run-time Processing

Average processing time of the various methods (mentioned earlier) for a 256x256 image on a Core 2 Duo 2.0 GHz computer, implemented using MATLAB is shown in Table X. Our method being HMM based takes little more computational time for detecting the curved texts and then verifying those. The other methods of Lie et al. [9], Cai et al. [20] and Wong et al. [14] do not perform well for detecting non-horizontal or curved text regions. To give an idea about the run-time comparison, they have been run on such type of image datasets as well. Average processing time of different methods on IIT-R text dataset is given in Table X. A comparative study of average run-time performance among different classifiers (mentioned in Table IX) is also provided in Table XI.

**Table X: Average Processing Time of different methods on IIT-R text dataset**

| Method | Horizontal Text (in Seconds) | Linear Non-Horizontal Text (in Seconds) | Curved Text (in Seconds) |
|---|---|---|---|
| Liu et al. [9] | 20.1 | 24.3 | 19.8 |
| Cai et al. [20] | 6.3 | 7.2 | 6.7 |
| Wong et al. [14] | 1.8 | 1.5 | 1.6 |
| Neumann et. al [37] | 1.4 | 1.7 | 1.6 |
| Shivakumara et al. [1] | 7.5 | 9.1 | 10.4 |
| **Proposed(using HMM)** | 9.2 | 10.9 | 12.2 |

**Table XI: Average Processing Time of our proposed method using different classifiers on IIT-R text dataset**

| Method | Horizontal Text (in Seconds) | Linear Non-Horizontal Text (in Seconds) | Curved Text (in Seconds) |
|---|---|---|---|
| SVM | 8.4 | 10.2 | 11.5 |
| AdaBoost | 8.8 | 10.5 | 11.9 |
| Random Forest | 8.2 | 9.9 | 11.3 |
| CRF | 9.1 | 11.2 | 12.4 |
| HMM | 9.2 | 10.9 | 12.2 |

### 4.8. Error Analysis

Although our method performs quite well in a large diversity of images with varying fonts, text-sizes or complex backgrounds, there are a few cases where our method fails to perform up to our degree of satisfaction. The proposed method might give contradictory results if there is a large difference among font sizes in the same image. Such a situation would make the value needed for the thickening of the resultant skeletons before they are sent for HMM verification, difficult, thus lowering the accuracy of the method. Such cases (Fig.22) are rare in general situations, as the font size difference in the same image is usually not extremely large.

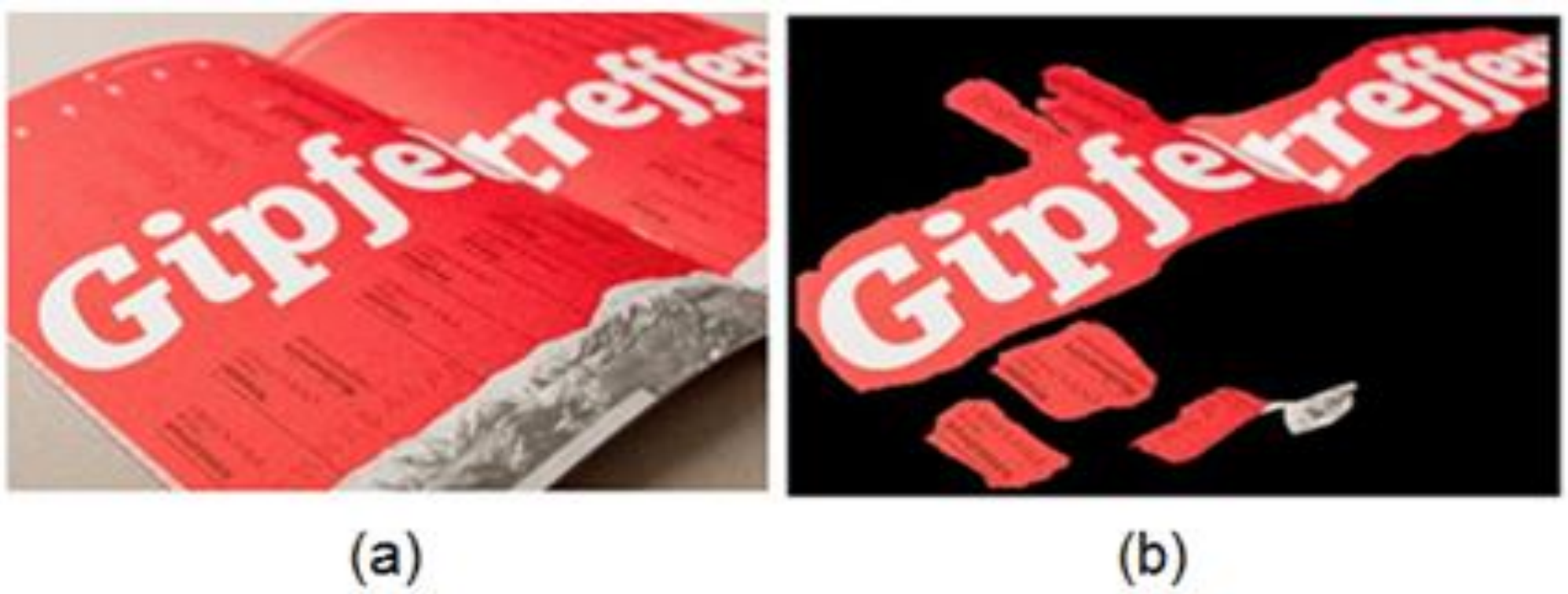

**Fig.22. False text detection (a) Original image (b) Text detection results by our system.**

## 5. Conclusion and Future Work

We have put forward a technique for text detection which is capable of handling text in video frames as well as scene images. In this paper, text orientation and text-shape are our main concern for the detection purpose. The Laplacian filtering in the frequency domain helps us in identifying the probable text regions, which are then skeletonized and segmented into disjoint connected components (CCs). These disjoint CCs are recombined into probable text shapes based on different possible combinations, which upon HMM verification, identifies horizontal, linear non-horizontal as well as curvilinear texts efficiently. In future we intend to deal with the shapes of the skeletal segments in further details, thus hoping to improve further on the efficiency and accuracy of our method.

There are not many efficient methods that perform well on detection of curvilinear texts. Most methods detect horizontally aligned texts. Some researchers have looked into linear non-horizontal orientation of texts, but curved text detection has not been efficiently explored into yet. Our method has shown promising results in detection of such curvedly oriented texts. From the results, it was noted that the proposed method performs well over a large range of images, for horizontal, linear non-horizontal and curved texts. Also, it performs better than the existing methods as far as non-horizontal or curvilinear text orientations are concerned. In the future, we plan to deal with the problems of false positives, and partially detected text lines, especially those belonging to curved text areas.